%% file: main.tex
\documentclass[11pt]{article}

\usepackage[final]{acl}
\usepackage{longtable}
\usepackage{times}
\usepackage{latexsym}
\usepackage{amsmath}
\usepackage{amssymb}
\usepackage{mathtools}
\usepackage{amsthm}
\usepackage{booktabs}
\usepackage{array}
\usepackage{pgfplots}
\pgfplotsset{compat=1.18}
\usepackage{listings}
\usepackage{multirow}
\usepackage{caption}
\usepackage{adjustbox}
\usepackage{tikz}
\usetikzlibrary{arrows.meta, positioning}

\definecolor{bluebg}{RGB}{228,240,255}
\definecolor{greenbg}{RGB}{229,246,235}
\definecolor{redbg}{RGB}{255,229,229}
\definecolor{yellowbg}{RGB}{255,246,214}
\definecolor{graybg}{RGB}{245,245,245}
\definecolor{darkblue}{RGB}{28,84,156}
\definecolor{darkgreen}{RGB}{39,130,72}
\definecolor{darkred}{RGB}{177,55,55}
\definecolor{darkorange}{RGB}{176,105,30}

\usepackage{xcolor}
\usepackage{enumitem}
\usepackage[table]{xcolor}
\definecolor{propcolor}{RGB}{225,235,255}
\definecolor{opensourcecolor}{RGB}{230,245,230}
\definecolor{baselinecolor}{RGB}{245,235,225}

\usepackage[T1]{fontenc}

\usepackage[utf8]{inputenc}

\usepackage{microtype}

\usepackage{inconsolata}

\usepackage{graphicx}
\usepackage{xcolor}

\definecolor{reliable}{RGB}{226,245,231}
\definecolor{potemkin}{RGB}{255,239,207}
\definecolor{shortcut}{RGB}{238,238,238}
\definecolor{failure}{RGB}{255,229,229}
\definecolor{darkred}{RGB}{160,40,40}
\definecolor{darkgreen}{RGB}{35,115,65}
\definecolor{darkorange}{RGB}{170,100,20}
\usepackage{xspace}

\title{Do Audio Language Models Use Paralinguistic Evidence?\\ Counterfactual Audits for Response Evaluation}

\author{
\textbf{Kevin Miller}\textsuperscript{*},  
\textbf{Arjun Chandra}\textsuperscript{*}\\
\textbf{Venkatesh Saligrama} \\
Boston University \\
\texttt{\{nivek, ac25, srv\}@bu.edu}
}

\begin{document}
\maketitle
\begingroup
\renewcommand\thefootnote{\*}
\footnotetext{
*Equal contribution. Correspondence to \texttt{srv@bu.edu}.\\
\hspace*{\parindent}Code/Data: \href{https://github.com/kjmillerCURIS/CounterfactualParalinguistics}{kjmillerCURIS/CounterfactualParalinguistics}
}
\endgroup







    

    


\input{sec/abstract}

\input{sec/intro}
\input{sec/related_main}
\input{sec/framework}
\input{sec/benchmark}

\input{sec/results}

\input{sec/conclusion}
\input{sec/acknowledgments}

\bibliography{custom}

\appendix
\input{sec/appendix}

\end{document}

%% file: sec/abstract.tex
\begin{abstract}
Audio-language models (ALMs) are increasingly used as judges for speech-to-speech systems, but a judge that receives audio may not actually use paralinguistic evidence. We introduce counterfactual audits for paralinguistic response evaluation. Each audit item holds the transcript fixed while varying affect, prosody, or the timing of an affective shift, forcing a valid judge to track the audio cue rather than lexical content or response style. We evaluate ALM judges using a native pairwise judgment task and a contrastive matching control, then further decompose each item into its constituent perception and response-mapping skills. Across Gemini, GPT, and open audio models, we find that contrastive success often overstates native judge reliability, and similar aggregate accuracies can hide different failure modes. These results suggest that current ALM judges require thorough behavioral audits before being used for automatic speech-to-speech evaluation.
\end{abstract}

%% file: sec/intro.tex
\section{Introduction}
\begin{figure*}[t]
\centering
\includegraphics[width=\textwidth]{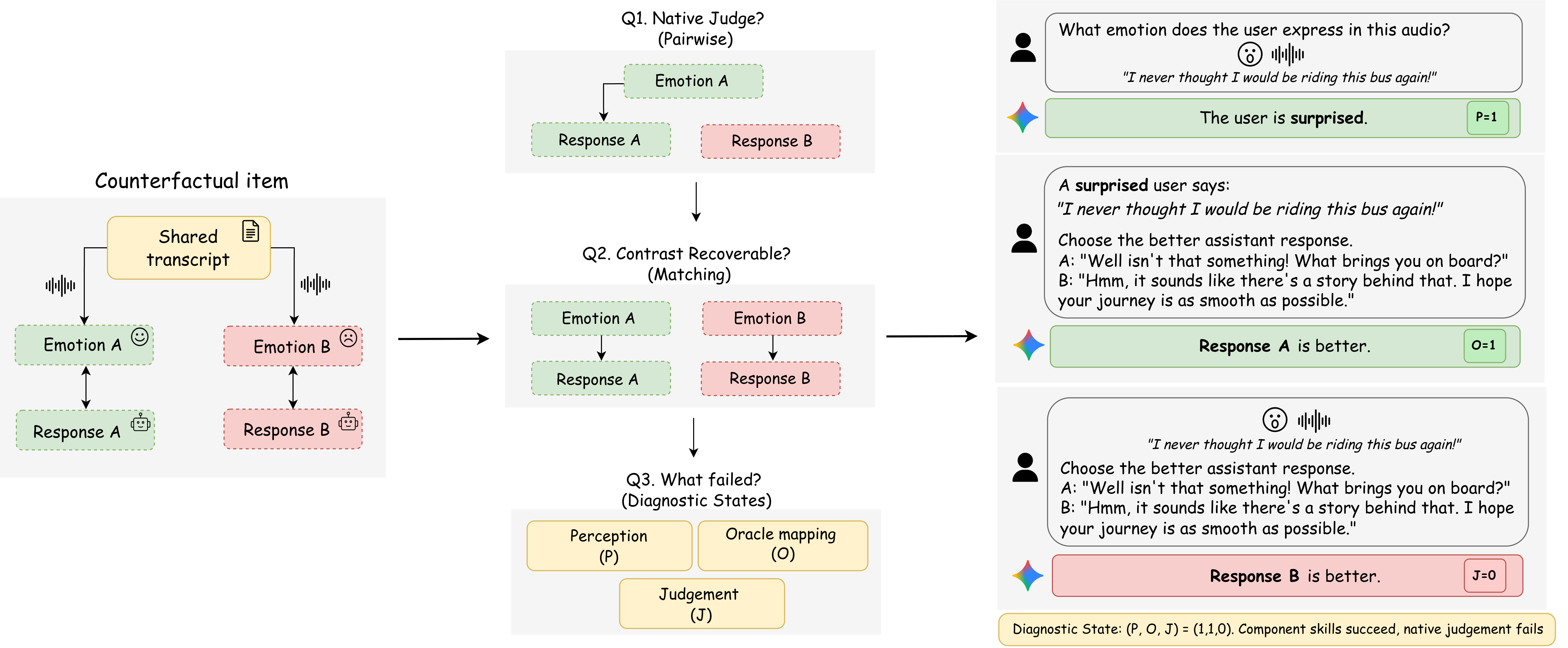}

\caption{
\textbf{Counterfactual audio-judge audit.} 
\textbf{Left:} 
The audit holds the user's words fixed while changing only the paralinguistic realization. \textbf{Middle:} \textsc{Pairwise} provides one audio context and two candidate responses. If this fails, \textsc{Matching} provides both
counterfactual audio contexts and responses. Component probes then test the constituent task skills to identify failure sources.
\textbf{Right:} The judge correctly identifies the paralinguistic state
(\(P=1\)) and maps that state to the better response when supplied in text
(\(O=1\)), but fails in the native audio judgment (\(J=0\)).
}
\vspace{-12pt}
\label{fig:diagnostic-cascade-example}
\end{figure*}

Paralinguistic cues such as tone, rhythm, pitch, and timing are central to spoken interaction
\citep{scherer2003vocal,crystal1975paralinguistics}. The same words can require different
assistant responses depending on how they are spoken: a user may sound surprised, hesitant,
impatient, or frustrated even when the transcript alone is ambiguous. As speech-to-speech
assistants become more capable, evaluation must therefore move beyond text. We need to know
whether a spoken response is appropriate not only for what the user said, but also for how the
user said it.

A natural way to scale this evaluation is to use audio-language models (ALMs) as automatic
judges \citep{manakul2025audiojudge,chiang2025audioaware,jiang2025s2sarena}. An ALM judge
can listen to an interaction and decide which assistant response is better. If reliable, such
judges could support benchmarking, reward modeling, deployment monitoring, and regression
testing for spoken assistants. However, this creates a second-order evaluation problem: before
using an ALM as a judge, we must audit whether the judge itself uses the relevant audio evidence.
A judge that relies mostly on transcripts, response style, or generic conversational preferences
is not a trustworthy evaluator of paralinguistics.

We study this problem through \emph{counterfactual audits}. Each audit item holds the user's
words fixed while changing only the paralinguistic realization. The candidate responses are then paired with each realization. When the relevant audio cue changes, a valid judge should change its response judgment accordingly. 

A failed counterfactual audit is underdetermined. If the judge selects the incorrect response, it does not reveal whether the judge failed to
hear the paralinguistic cue, map the cue to the right response, or compose
these skills in the final judgment. We therefore employ the audit framework shown in
Figure~\ref{fig:diagnostic-cascade-example}.

\noindent\textbf{Native judgment.}
The target setting is \textsc{Pairwise}, in which the judge receives one user input
and two candidate responses, and must choose the response appropriate for that
input. We call this judgment native because it is the closest to a standard setting for using an ALM as an automatic evaluator \cite{manakul2025audiojudge}.

\noindent\textbf{Contrastive recoverability.}
If native judgment fails, we test whether performance is recoverable when the paralinguistic contrast is made explicit. In \textsc{Matching}, the judge receives both
audio realizations and both candidate responses, and must determine the correct matching \cite{zhu2026testtime}. 
If \textsc{Matching} succeeds, the explicit contrast 
forces the model to look past the identical transcripts and rely on the audio cue. 
This reveals that the model has the requisite paralinguistic judgment ability but fails to deploy it in the native evaluation setting, making \textsc{Matching} an effective diagnostic tool.
%

\noindent\textbf{Component attribution.} In addition to the end-to-end judgment (\(J\)) task, we identify sources of model failures by implementing a separate perception (\(P\)) and 
oracle response mapping (\(O\)) probe for each item. The perception probe
asks whether the judge correctly identifies the user's paralinguistic state. The
response-mapping probe asks whether the judge can choose the correct
response when that user's paralinguistic state is provided. Together, the state
\((P,O, J)\) separates heterogeneous failure modes across different judge models.

We apply this audit framework to both single-turn and multi-turn settings. Single-turn examples isolate static
paralinguistic response selection: given one utterance rendered with different affective states, does
the judge select the response that fits the spoken state? The multi-turn examples then add temporal
and causal structure in which the transcript is fixed, but the timing of the user's affective shift changes,
so the correct response depends on when the user became frustrated and what caused the shift. This
captures a deployment-relevant difficulty for spoken assistants, where emotion is often interpreted
through dialogue history \citep{poria2019meld,du2025mtalkbenchevaluatingspeechtospeechmodels,arora2025talking}.

Our experiments show that current ALM judges are fragile in distinct ways. Some models recover their performance in the \textsc{Matching} task but degrade sharply in \textsc{Pairwise}, showing
that contrastive success does not imply native judge reliability. The component probes further demonstrate
that aggregate accuracy hides heterogeneous failure modes across judges. In multi-turn examples, native pairwise evaluation remains especially brittle, indicating that temporal-causal paralinguistic judgment remains challenging.

\noindent \textbf{Contributions.} First, we introduce counterfactual audits for ALMs used as speech-to-speech judges. Second, we propose a diagnostic framework that
combines native pairwise judgment, a contrastive matching control, and component probes. Third, we present an empirical audit of Gemini, GPT, and open audio models, showing that aggregate accuracy does not certify ALM judge reliability.

%% file: sec/related_main.tex
\section{Related Work}
Large Language Models (LLMs) are routinely used as text evaluators, but they can exhibit systematic biases \citep{dubois2024lengthcontrolled,zheng2023judging} and display ``Potemkin understanding'' \citep{mancoridis2025potemkin}, motivating decomposition-based reliability improvements \citep{lee2025checkeval,li2025dnaeval}. Concurrently, audio-capable models are increasingly deployed as automatic evaluators \citep{manakul2025audiojudge,chiang2025audioaware} and reward models \citep{ji2025wavreward,ge2026sagelm, yang2026paras2sbenchmarkingaligningspoken} in speech-to-speech settings. These applications typically assume that access to the audio modality implies reliable paralinguistic reasoning---an assumption challenged by recent findings \citep{chandra2026trace,chen-etal-2026-audio}. While existing benchmarks evaluate paralinguistic instruction following \citep{jiang2025s2sarena,held2025cava} or emotion perception \citep{yang2021superb, huang2024dynamic}, they provide limited diagnostic leverage for understanding how judges use these cues in downstream preference decisions. We bridge this gap by explicitly isolating paralinguistic reasoning in audio evaluators. Extended discussion of related evaluation frameworks is provided in Appendix \ref{sec:related}.

%% file: sec/framework.tex
\begin{table*}[t]
\centering
\scriptsize
\resizebox{\linewidth}{!}{%
\begin{tabular}{cccp{3.3cm}p{8.8cm}}
\toprule
$P$ & $O$ & $J$ & Diagnostic State & Interpretation \\
\midrule
\rowcolor{failure}0 & 0 & 0 & Full-stack failure & Neither component probe succeeds; final judgment fails. \\
\rowcolor{failure}0 & 1 & 0 & Perception bottleneck & Response-mapping succeeds, but the judge does not perceive the paralinguistics. \\
\rowcolor{failure}1 & 0 & 0 & Response-mapping bottleneck & The judge perceives the paralinguistics, but cannot map it to the correct response. \\
\rowcolor{potemkin}1 & 1 & 0 & Potemkin / orchestration failure & The judge passes perception and response-mapping, but final judgment still fails. \\
\rowcolor{shortcut}0 & 0 & 1 & Accidental success & The final judgment is correct despite both component probes failing. \\
\rowcolor{shortcut}0 & 1 & 1 & Shortcut: perception & The final judgment is correct despite perception probe failing. \\
\rowcolor{shortcut}1 & 0 & 1 & Shortcut: response-mapping & The final judgment is correct despite response-mapping probe failing. \\
\rowcolor{reliable}1 & 1 & 1 & Reliable integrated judgment & Perception, response-mapping, and final judgment all succeed. \\
\bottomrule
\end{tabular}}
\caption{\textbf{Diagnostic truth table for ALM judge reliability.} Each row in the table indicates a different interpretation of the model's behavior. The final row is the reliable judge state, and the highlighted Potemkin row is a key  failure mode we uncover.}
\label{tab:truth-def}
\end{table*}

\section{Problem Setup and Framework}
\label{sec:framework}

Our goal is to audit audio-language models (ALMs) when they are used as judges of spoken interactions. The target use case is multi-turn spoken interaction, where an assistant response must fit not only the user's words but also the user's tone, affect, and the timing of an affective shift. This target is realistic, but a wrong judgment is hard to interpret. The judge may fail to hear the cue, fail to map the cue to the right response, recover the cue only when the contrast is explicit, or fail because temporal-causal reasoning is required. The framework below makes these different possibilities identifiable.

\subsection{Counterfactual audit items}
\label{subsec:counterfactual_examples}

Each audit item $E$ is a controlled counterfactual: 
\[
E=(C,\mathcal{A}^{0},\mathcal{A}^{1},R^{0},R^{1}),
\]
where $C$ is a shared transcript or conversation history, $\mathcal{A}^{0}$ and $\mathcal{A}^{1}$ are two audio realizations of the same words, and $R^{0}$ and $R^{1}$ are the responses appropriate for the two realizations. Importantly, the lexical content is fixed, and only the paralinguistic realization changes between $\mathcal{A}^{0}$ and $\mathcal{A}^{1}$.

This design gives the audit its validity. A text-only or lexically biased judge cannot solve the item consistently, because both branches of the counterfactual share the same transcript. A judge sensitive to paralinguistics should change its decision when the relevant audio cue changes. The audit therefore tests whether the judge measures the audio cue, rather than response style or lexical shortcuts.

\subsection{Native judgment and contrastive recoverability}
\label{subsec:protocols}

The native judgment setting is \textsc{Pairwise}. The judge receives one audio context $\mathcal{A}^{y}$ and two candidate responses $\{R^{0},R^{1}\}$, and must choose the response appropriate for that audio. 

In addition, we probe ALMs using a \textsc{Matching} setting as a contrastive recoverability control. In \textsc{Matching}, the judge receives both audio contexts $\mathcal{A}^{0},\mathcal{A}^{1}$ and both responses $R^{0},R^{1}$, and must match each response to the corresponding context. This setting is a control to test whether ALMs can distinguish the two counterfactual audio-response pairs when both alternatives are visible, but it is not a realistic deployment setting. Instead, a large \textsc{Matching}--\textsc{Pairwise} performance gap indicates a protocol dependence where the judge possesses the paralinguistic evaluation ability, but does not reliably deploy it in the standard \textsc{Pairwise} setting.

We also vary the amount of cueing in the prompt. A no-cue prompt asks for a direct judgment. A hard cue instructs the judge to focus on the user's emotion and prosody. For multi-turn examples, the transition cue tells the judge to track where the user's affect changes before choosing the response. Prompt cueing provides task-specific information, so a judge whose performance depends sharply on it may be useful in a scaffolded pipeline, but should not be treated as a plug-and-play evaluator.

\subsection{Component attribution: perception, mapping, and judgment}
\label{sec:framework_diagnostics}

Protocol comparisons tell us whether the native judgment is reliable, but they do not identify where the judge failed. We therefore evaluate each item through two component probes in addition to the native judgment task (Fig.~\ref{fig:diagnostic-cascade-example}):
\begin{align*}
P_i &= \mathbf{1}\{\text{perception probe is correct}\},\\
O_i &= \mathbf{1}\{\text{oracle response-mapping probe is correct}\},\\
J_i &= \mathbf{1}\{\text{native audio judgment is correct}\}.
\end{align*}
The perception probe $P_i$ tests whether the judge identifies the task-relevant paralinguistic state. For single-turn examples, this is the user's emotion or prosodic state. For multi-turn examples, this is the full affective trajectory (i.e., the user's emotion at every turn). The oracle response-mapping probe $O_i$ removes the audio perception burden by supplying the relevant state in text, and tests whether the judge can map that state to the appropriate response. The native judgment $J_i$ records whether the judge succeeds in the original end-to-end response evaluation task.

Each item is assigned a diagnostic state:
\[
z_i=(P_i,O_i,J_i)\in\{0,1\}^3
\]
and we report the empirical state distribution:
\[
\pi_{poj}=\frac{1}{n}\sum_{i=1}^{n}\mathbf{1}\{P_i=p,\,O_i=o,\,J_i=j\}.
\]
We use $(P,O,J)$ order throughout the paper. The reliable integrated state is $\pi_{111}$. A key  failure mode we uncover is $\pi_{110}$ where the judge succeeds on perception and response mapping, but fails when those abilities must be composed in the native audio judgment. We call this a Potemkin failure~\citep{mancoridis2025potemkin}. Other state groups distinguish perception bottlenecks, response-mapping bottlenecks, shortcut-like successes, and full-stack failures. The full eight-state table is shown in Table~\ref{tab:truth-def}.

\subsection{Task complexity: static versus temporal-causal judgment}
\label{subsec:interaction_complexity}

The final diagnostic axis is interaction complexity. Multi-turn spoken interaction is the deployment-relevant target, but it entangles perception, timing, causal attribution, and response mapping. We therefore use \textsc{single-turn-emotions} as a baseline task: it removes dialogue history and temporal reasoning, and tests whether the judge can use a static paralinguistic cue at all.

We then return to the multi-turn target through the \textsc{positional-emotion} task. These examples preserve counterfactual control while restoring temporal structure. The transcript is controlled, but the affective shift occurs at different turns. The correct response depends on when the user became frustrated and which assistant action caused the shift. Thus, a model that succeeds on single-turn examples but fails on positional examples is likely brittle to temporal-causal interaction structure.



%% file: sec/benchmark.tex
\section{Task Construction}
\label{sec:benchmark_construction}

The tasks instantiate the audit design in Section~\ref{sec:framework}. Each task removes or restores a specific source of difficulty. \textsc{single-turn-emotions} removes dialogue history and tests static paralinguistic response selection. \textsc{positional-emotion} restores temporal-causal structure while preserving counterfactual control. We also report \textsc{emotional-conversations}, an earlier prototype, in Appendix \ref{subsec:app_full_extended_results} because many of those examples can be solved from the final user turn alone.

\begin{figure*}[t]
\centering
\includegraphics[width=\textwidth]{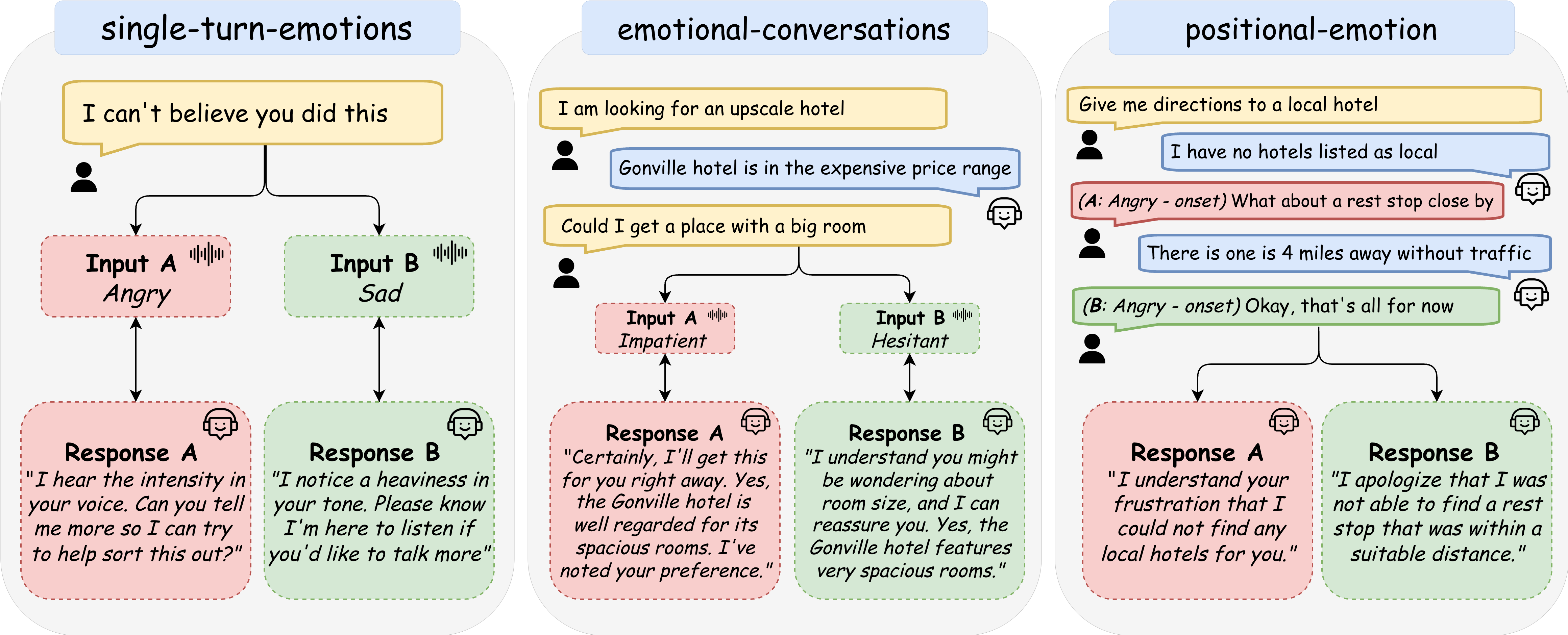}
\caption{\textbf{Task families in the audit.} \textsc{single-turn-emotions} isolates static paralinguistic response selection. \textsc{emotional-conversations} is an earlier multi-turn prototype reported in the appendix. \textsc{positional-emotion} has the timing and cause of the user's affective shift determine the correct response.}
\label{fig:benchmark-main}
\end{figure*}

\paragraph{Controlled synthesis.}
We use text-to-speech as an experimental control. The transcript, conversation history, and candidate responses are fixed while the paralinguistic realization changes. This gives precise counterfactual control so that if the judge's decision changes, the change can be attributed to the audio cue rather than lexical or contextual differences. We later demonstrate generalization of our findings to real speech in Section~\ref{sec:validity_ablation_main}.

\subsection{Single-turn task: static baseline}
\label{subsec:single_turn}

The \textsc{single-turn-emotions} task asks whether a judge can use a static paralinguistic cue before dialogue history, temporal localization, or causal attribution enter. Each item contains one user utterance rendered with two contrasting emotions, together with two candidate assistant responses. The correct response depends on the user's tone, not on the transcript alone.

\paragraph{Source data.}
We source lexical user inputs from the EmoCF subset of CAVA~\citep{held2025cava}. EmoCF contains lexically neutral user inputs grounded in diverse social situations from NormBank~\citep{ziems-etal-2023-normbank}. This makes it useful for response-level paralinguistic evaluation since the judge must decide which assistant response fits the spoken cue rather than rely on the words alone. Standard acted corpora such as CREMA-D and RAVDESS~\citep{cao2014cremad,ravdess} provide speaker diversity, but use a small set of repeated utterances; this is less suitable for our setting because response appropriateness must be evaluated across diverse lexical contexts.

\paragraph{Construction.}
We select 189 user inputs, which yields 189 items for the matching setting and 378 for the pairwise setting. Each input is paired with two contrasting target emotions, such as anger versus sadness or surprise versus neutrality. We synthesize user audios with \texttt{gpt-4o-mini-tts}, generate emotion-appropriate assistant responses with \texttt{gemini-2.5-flash}, and synthesize those responses with \texttt{gemini-2.5-flash-preview-tts}. The responses imply the relevant emotional adaptation without explicitly naming the emotion. Full prompts and construction details are provided in Appendix~\ref{app:single_turn_benchmark_construction}.


\begin{table}[t]
\centering
\footnotesize
\setlength{\tabcolsep}{3pt}
\renewcommand{\arraystretch}{1.1}

\begin{tabular}{lccc}
\toprule
 Statistic & \texttt{single-turn} & \texttt{emo-conv} & \texttt{pos-emo} \\
\midrule
\# Items   & 378 & 500 & 1000 \\
\# Emotions    & 4   & 8 & 7 \\
Avg turns      & 1.0 & 5.8 & 9.9 \\
Avg sec/utterance   & 5.5 & 5.2 & 4.9 \\
Avg words/utterance & 14.0 & 12.8 & 12.4 \\
\bottomrule
\end{tabular}

\caption{\textbf{Key statistics of each audit task.}}
\label{tab:dataset_stats}
\end{table}

\subsection{Positional multi-turn task: temporal-causal judgment}
\label{subsec:multi_turn_benchmark_construction}

The \textsc{positional-emotion} task tests a harder form of paralinguistic judgment. A user may become frustrated only after an earlier goal is not met, and the appropriate response may depend on identifying which event caused the change. The task therefore asks whether a judge can localize an affective shift in a conversation and use that causal interpretation to select the right response.

\paragraph{Source data.}
We seed the task with 500 conversations from Open Dialogue (OD3)~\citep{chan2023domain}. OD3 aggregates task-oriented dialogue sources including KVRET~\citep{kvret}, MultiWOZ~\citep{multiwoz}, DSTC11~\citep{zhao2023dstc11}, NOESIS-II~\citep{noesisii}, and SIMMC-2.1~\citep{kottur-etal-2021-simmc}. These sources cover assistant-like interactions such as travel, booking, shopping, navigation, reminders, and other goal-directed tasks. 

\paragraph{Goal-failure injection.}
Given a source conversation $S$, we identify user goals and create two counterfactual branches. In one branch, the assistant fails to satisfy goal $g_a$ at turn $t_a$; in the other, it fails to satisfy goal $g_b$ at turn $t_b$, with $t_a\neq t_b$. The final transcript is controlled across the pair, but the onset of negative affect occurs at a different position. The candidate responses are written so that each response addresses the corresponding cause of the affective shift. Source conversations are rejected when a plausible controlled contrast cannot be created.

\paragraph{Audio rendering and quality control.}
We render turns with \texttt{gpt-4o-mini-tts}. For paralinguistically important turns, we generate multiple attempts and keep the best accepted rendering. Acceptance uses \texttt{emotion2vec} scores~\citep{ma2023emotion2vec} and acoustic primitives such as pitch and speech rate. Full rendering prompts, filtering criteria, and quality-control details are provided in Appendix~\ref{app:audio_rendering_and_qc}.

\input{sec/human_accuracy_tables}

\paragraph{Perception target scoring.}
For \textsc{single-turn-emotions}, the perception probe asks for the user's emotion. For \textsc{positional-emotion}, the judge must correctly recognize the full affective trajectory (i.e., the user's emotion at every turn) for the $(P,O,J)$ state analysis. This keeps the positional multi-turn diagnostic states comparable to the single-turn diagnostics while preserving the temporal requirement.


\subsection{Human validation}
\label{subsec:human_validation_main}
\input{sec/human_validation_new}

%% file: sec/human_accuracy_tables.tex


\begin{table}[t]
\centering
\setlength{\tabcolsep}{3pt}
\renewcommand{\arraystretch}{1.1}

\begin{tabular}{lccc}
\toprule
Subject & $N$ & Accuracy (\%) & 95\% CI \\
\midrule
Subject A & 50 & 66.0  & [52.2, 77.6] \\
Subject B & 50 & 54.0  & [40.4, 67.0] \\
Subject E & 50 & 64.0  & [50.1, 75.9] \\
Subject G & 50 & 60.0  & [46.2, 72.4] \\
Subject H & 50 & 62.0  & [48.2, 74.1] \\
Subject I & 50 & 72.0  & [58.3, 82.5] \\
Subject J & 50 & 58.0  & [44.2, 70.6] \\
Subject K$^\dagger$ & 25 & 100.0 & [86.7, 100.0] \\
\bottomrule
\end{tabular}

\caption{\textbf{Single-turn human validation.} Human judge accuracy on the single-turn-emotions task, with 95\% Wilson confidence intervals. $^\dagger$ denotes an author of this paper.}
\label{tab:human-acc-single-turn-emotion}
\end{table}

\begin{table}[t]
\centering
\setlength{\tabcolsep}{3pt}
\renewcommand{\arraystretch}{1.1}

\begin{tabular}{lccc}
\toprule
Subject & $N$ & Accuracy (\%) & 95\% CI \\
\midrule
Subject C & 50 & 90.0 & [78.6, 95.7] \\
Subject D & 49 & 93.9 & [83.5, 97.9] \\
Subject F & 50 & 92.0 & [81.2, 96.8] \\
Subject I & 26 & 88.5 & [71.0, 96.0] \\
Subject L$^\dagger$ & 40 & 87.5 & [73.9, 94.5] \\
\bottomrule
\end{tabular}

\caption{\textbf{Multi-turn human validation.} Human judge accuracy on the positional-emotion task, with 95\% Wilson confidence intervals. $^\dagger$ denotes an author of this paper.}
\label{tab:human-acc-positional-emotion}
\end{table}

\begin{table*}[t]
\centering
\small
\resizebox{\textwidth}{!}{%
\begin{tabular}{lcccccccc}
\toprule
& \multicolumn{4}{c}{\textbf{Single-turn-emotions}}
& \multicolumn{4}{c}{\textbf{Positional-emotion}} \\
\cmidrule(lr){2-5}\cmidrule(lr){6-9}
& \multicolumn{2}{c}{\textsc{Pairwise}}
& \multicolumn{2}{c}{\textsc{Matching}}
& \multicolumn{2}{c}{\textsc{Pairwise}}
& \multicolumn{2}{c}{\textsc{Matching}} \\
\cmidrule(lr){2-3}\cmidrule(lr){4-5}\cmidrule(lr){6-7}\cmidrule(lr){8-9}
\textbf{Judge} & No cue & Hard cue & No cue & Hard cue & No cue & Trans. cue & No cue & Trans. cue \\
\midrule

\textbf{Proprietary models} & & & & & & & & \\

Gemini-2.5-Flash & 53.1 & 59.0 & 74.3 & 74.0 & 48.7 & 53.9 & 49.0 & 62.0 \\
Gemini-2.5-Pro   & 52.1 & 60.1 & 80.6 & 86.0 & 50.2 & \textbf{54.5} & \textbf{57.6} & 65.8 \\
Gemini-3-Flash   & 52.4 & 58.7 & 79.4 & 83.1 & 50.3 & 53.0 & 53.6 & \textbf{79.4} \\
Gemini-3-Pro     & \textbf{55.2} & \textbf{65.3} & \textbf{84.3} & \textbf{91.0} & 48.7 & 51.6 & 55.0 & 66.4 \\
GPT-4o-mini      & 52.3 & 55.4 & 45.3 & 54.9 & \textbf{52.9} & 52.0 & 49.2 & 52.0 \\
GPT-4o           & 54.2 & 53.8 & 57.1 & 55.9 & 50.3 & 49.2 & 46.7 & 51.6 \\
Nova-2-Pro       & 48.4 & 58.4 & 55.1 & 53.6 & 50.7 & 48.3 & 49.6 & 50.5 \\

\midrule

\textbf{Open-source models} & & & & & & & & \\

Phi-4-Multimodal-6B & 53.6 & 48.7 & 49.2 & 49.2 & 51.6 & 49.3 & 45.2 & 47.0 \\
Qwen2.5-Omni-7B     & 50.0 & 47.1 & 51.7 & 54.1 & 47.3 & 51.1 & 47.1 & 48.4 \\
DeSTA2.5-Audio-8B   & 48.9 & 47.9 & 49.2 & 46.1 & 50.6 & 50.3 & -- & -- \\
Voxtral-Small-24B   & 52.9 & 53.2 & 53.9 & 54.0 & 49.7 & 49.7 & 48.5 & 47.2 \\

\bottomrule
\end{tabular}}
\vspace{-4pt}
\caption{
\textbf{Headline end-to-end accuracy (\%).} Gemini models perform well in the \textsc{Matching} setting but collapse toward chance in \textsc{Pairwise}. Nearly all other judges remain near chance. Full Wilson confidence intervals for individual accuracies and paired
bootstrap confidence intervals for protocol gaps are reported in Appendix~\ref{subsec:app_CIs}.
}
\vspace{-12pt}
\label{tab:headline-results}
\end{table*}

%% file: sec/human_validation_new.tex
The labels for each item are determined by the counterfactual construction. We therefore use human validation to confirm that the paralinguistics are perceivable by listeners and that the correct response choices are sufficiently specified.

\noindent\textbf{Setup.} We recruited ten students from our home institution (independent of this paper) to serve as annotators for the \textsc{single-turn-emotions} and \textsc{positional-emotion} tasks. Annotators were a mix of native English speakers and non-native speakers with demonstrated English proficiency. The recruitment process was informal. All annotators were made aware of how their ratings would be used and consented to their use in this work and their public release. Two authors from this paper served as additional annotators but did not annotate any task that they themselves had authored. Seven independent students and one author annotated subsets of \textsc{single-turn-emotions}, while four independent students and one author annotated subsets of \textsc{positional-emotion} (one of the independent students annotated subsets from both tasks). Most annotators annotated 50 items, but some annotated as few as 25.

Annotators sampled items in the \textsc{Pairwise} format. For \textsc{single-turn-emotions}, they were given the ``hard'' cue from the ALM judgment experiments, and for \textsc{positional-emotion} they were given the ``transition'' cue. Items were presented through the browser interfaces shown in Figs.~\ref{fig:appendix_humeval_single} and \ref{fig:appendix_humeval_multi}.

Recent work on human annotation suggests that disagreement in subjective or interpretive tasks should not be treated as mere noise, but can reflect meaningful variation in annotator
perspective, task difficulty, or background knowledge \citep{uma2021learning,basile-etal-2021-need,davani2022dealing,huang2026deepfact}. For that reason, we report accuracy (agreement with ground truth) for each human annotator individually.

\noindent\textbf{Results.} Tabs.~\ref{tab:human-acc-single-turn-emotion} and \ref{tab:human-acc-positional-emotion} show the accuracy achieved by each human subject on the two tasks, with 95\% Wilson confidence intervals. Human accuracy on \textsc{single-turn-emotions} ranges from 54.0--100.0\%, with a median of 63.0\%, suggesting that paralinguistic judgments depend on listener sensitivity, language background, and attention to subtle prosodic cues. Meanwhile, human accuracy on \textsc{positional-emotion} ranges from 87.5--93.9\%, with a median of 90.0\%, suggesting that the multi-turn task is uniformly easy for humans.

There are several factors that may account for the  difference in human performance between the two tasks. First, a multi-turn context provides more information and thus a stronger signal for any judge who is capable of extracting it. Second, even when the \textsc{Pairwise} format is used, the multi-turn task is still a paralinguistic \textit{comparison} task, as it relies on a shift in emotion, while the single-turn task relies on classifying an emotion in isolation. And finally, there may be differences in the distribution of paralinguistics and semantics themselves. We leave it to future work to explore the relative difficulty of different paralinguistic tasks for both humans and machines.


%% file: sec/results.tex
\section{Experiments and Results}
\label{sec:experiments}

We organize the experiments as an audit of each judge. Each result answers one question in the diagnostic cascade shown in the middle panel of
Figure~\ref{fig:diagnostic-cascade-example}. We evaluate a range of frontier proprietary and open-source ALM judges on the two main task families in our audit introduced in Section  \ref{subsec:interaction_complexity}. Full prompts and implementation details are in Appendix \ref{subsec:app_judge_prompts}.

\begin{figure*}[t]
\centering
\includegraphics[width=\textwidth]{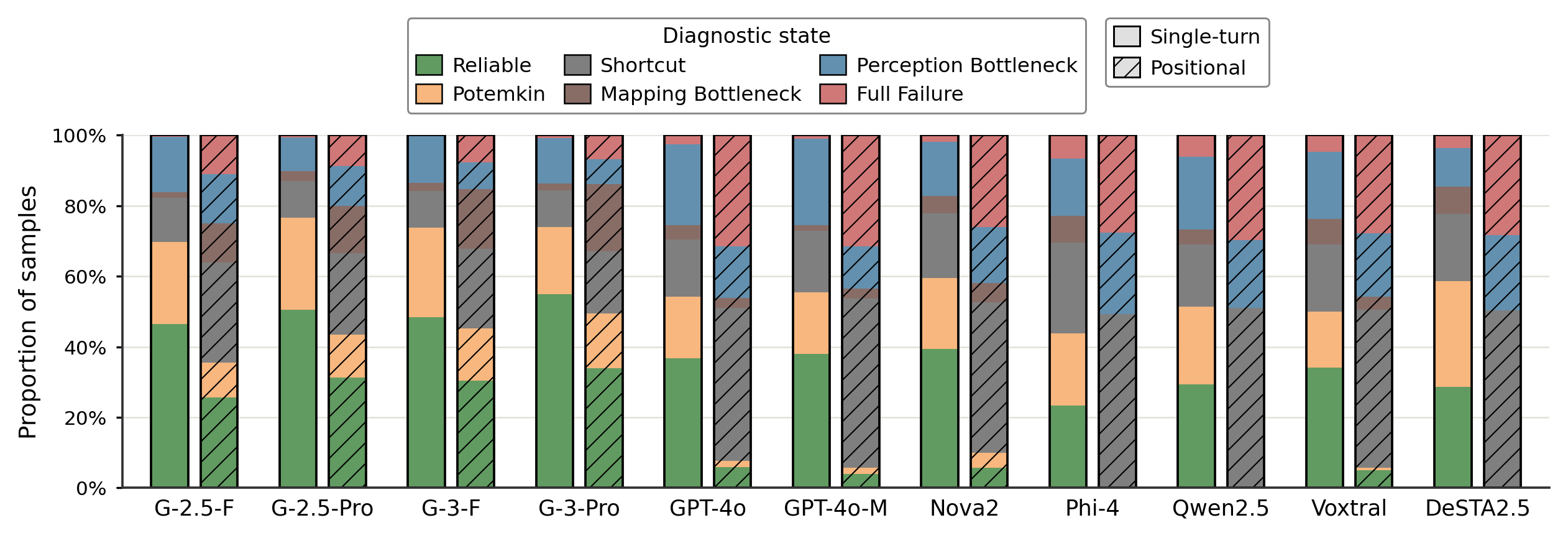}
\vspace{-25pt}
\caption{
\textbf{Judge state distributions.} Colors group the \((P,O,J)\) states into reliable integrated judgment, component
bottlenecks, Potemkin failures, shortcut-like successes, and full-stack
failures.}
\label{fig:state-distribution}
\end{figure*}

\subsection{Overall task performance and protocol collapse}
\label{sec:protocol-reliability}

Table~\ref{tab:headline-results} gives the headline end-to-end accuracies. We
read this table through the first two questions in the cascade. The
\textsc{Pairwise} columns show whether the judge works in the native pairwise judgment setting. The \textsc{Matching} columns test whether access to both counterfactual audio realizations and responses can recover performance in a matching setting when native judging fails.

\noindent\textbf{How does ALM performance compare to humans?} In the native \textsc{Pairwise} judgment setting, the strongest ALM is roughly on par with the median human judge on \textsc{single-turn-emotions}, while on \textsc{positional-emotion} there is a clear gap between ALM and human performance. In particular, note from Tab.~\ref{tab:ci_main_positional} (specifically the \textsc{Pairwise} rows) that all ALM confidence intervals lie entirely below 57.6\%, while every human subject's confidence interval (Tab.~\ref{tab:human-acc-positional-emotion}) lies entirely above 71.0\%, with the median lower bound at 78.6\%. Thus, we can say with statistical certainty that ALM judges are underperforming human judges on multi-turn positional paralinguistic reasoning.

\noindent\textbf{Protocol collapse.} The single-turn task already shows a large gap between the \textsc{matching} setting and
native \textsc{pairwise} setting. Gemini models often solve the contrastive matching problem, but
they are much less reliable when asked to judge one user audio context at a time. For
example, Gemini-3-Pro reaches $91.0\%$ in \textsc{Matching} with the hard cue,
but only $65.3\%$ in \textsc{Pairwise}; Gemini-2.5-Pro drops from
$86.0\%$ to $60.1\%$. 

\begin{figure}[t]
    \centering
    \includegraphics[width=0.49\textwidth]{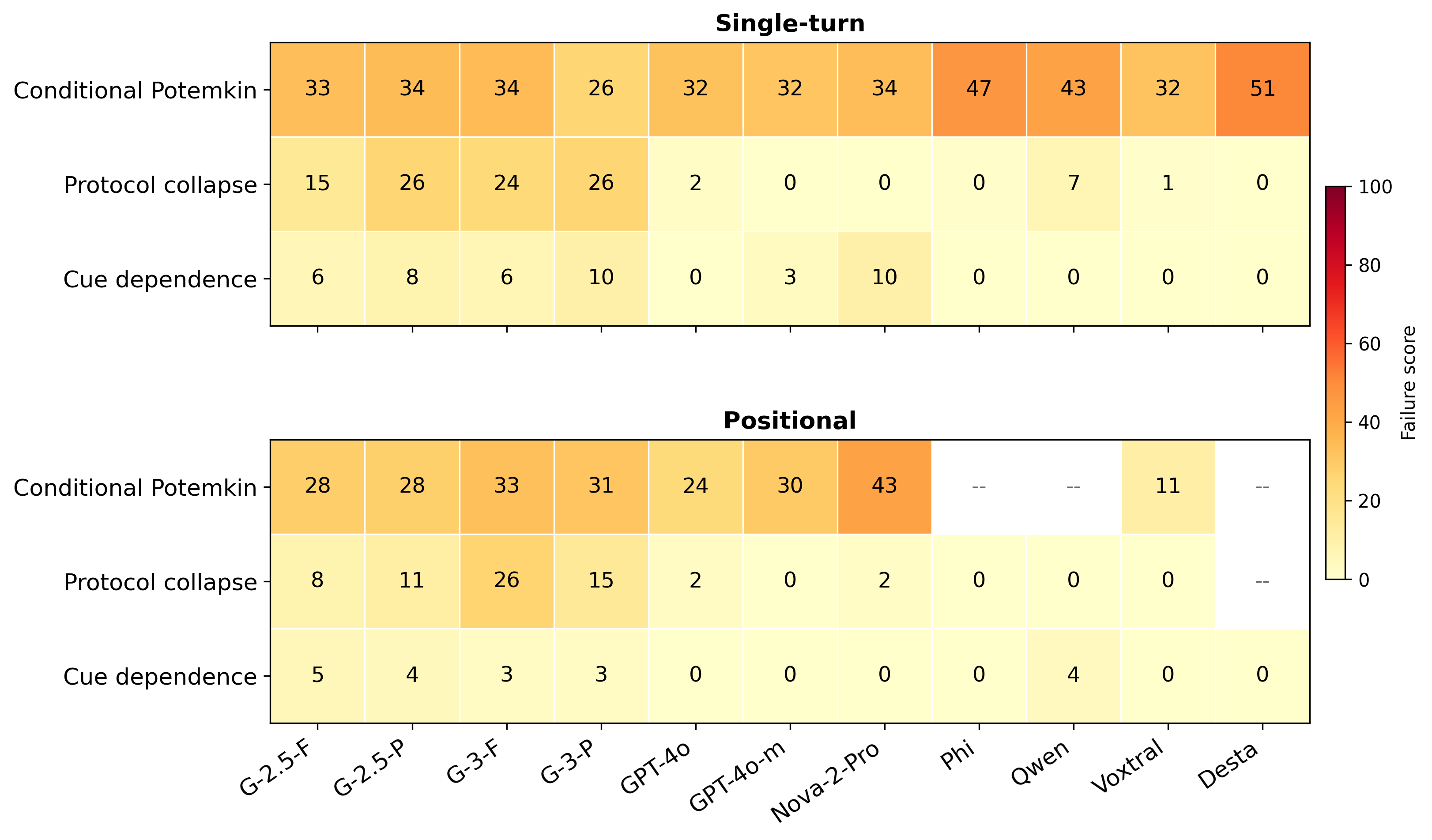} 
    \vspace{-20pt}
    \caption{\textbf{Failure heatmaps.} We quantify additional failure modes with conditional Potemkin rates, protocol collapse, and cue dependence.}
    \label{fig:additional_probes}
\end{figure}

The positional task makes the failure sharper. With the transition cue,
Gemini-3-Flash reaches $79.4\%$ in \textsc{Matching}, showing that some judges
can solve the matching problem when the paralinguistic contrast and task schema are
explicit. However, the same model reaches only $53.0\%$ in
\textsc{Pairwise}; Gemini-3-Pro reaches $51.6\%$, GPT-4o reaches
$49.2\%$, and Qwen-2.5-Omni-7B reaches $51.1\%$. Thus, contrastive success does not translate to judge reliability. A model may distinguish the two
counterfactuals when both are shown, but fail when the same paralinguistic cue must
control a single-context decision. We quantify this protocol collapse for each model in Fig.~\ref{fig:additional_probes} as the difference in accuracy between the \textsc{Matching} and \textsc{Pairwise} settings, and we report paired bootstrap confidence intervals in Appendix \ref{subsec:app_CIs}.



\subsection{What failed? Component attribution}
\label{sec:component-attribution}

The protocol collapse demonstrates that ALMs rely on explicit contrastive formats. However, it does not explain why many of the proprietary and open-source judges perform near chance on both tasks. To understand why they fail—specifically, whether the bottleneck lies in perceiving the audio cue, mapping it to a response, or composing these skills in a native judgment—we must examine the item-level diagnostic states defined in Section~\ref{sec:framework_diagnostics}.

\begin{table*}[t]
\centering
\small
\caption{
\textbf{Response quality balance check.} Absolute differences $|R^0 - R^1|$ across counterfactual response pairs in \textsc{Single-turn-emotions} confirm that irrelevant lexical and acoustic factors do not bias the judges.}
\begin{tabular}{lcccccccc}
\toprule
\textbf{Statistic}
& \multicolumn{4}{c}{\textbf{Audio Quality (DNSMOS, 1--5)}}
& \multicolumn{4}{c}{\textbf{Lexical Quality (LLM-Judge, 1--5)}} \\
\cmidrule(lr){2-5}\cmidrule(lr){6-9}
& SIG & BAK & OVRL & P808 & Helpfulness & Relevance & Specificity & Naturalness \\
\midrule
Mean Difference  & 0.14 & 0.15 & 0.19 & 0.20 & 0.41 & 0.74 & 0.51 & 0.69 \\
Median Difference  & 0.10 & 0.10 & 0.13 & 0.18 & 0.00    & 1.00    & 0.00    & 1.00    \\
Std Dev Difference & 0.13 & 0.17 & 0.17 & 0.16 & 0.63 & 0.76 & 0.61 & 0.72 \\

\bottomrule
\end{tabular}

\label{tab:response-balance-check}
\end{table*}

Figure~\ref{fig:state-distribution} shows the aggregated state distribution for each judge in each task. We focus on diagnosing failures in the standard \textsc{Pairwise} setting. Each bar decomposes evaluated examples into reliable integrated judgments, Potemkin failures, component bottlenecks, shortcuts, and full-stack failures. 

Crucially, two judges with similar end-to-end accuracy can place their probability mass in very different failure states. In the single-turn task, the strongest Gemini models are clearly able to reason from emotion as their oracle response-mapping ability is high, but their
native audio judgments often fail. Gemini-2.5-Pro has $26\%$ Potemkin mass,
Gemini-3-Flash has $25\%$, and Gemini-3-Pro has $19\%$. Conditioned on both
component probes succeeding, Gemini-3-Pro still fails the native audio judgment
$26\%$ of the time, and we report these full ``Conditional Potemkin'' rates across all models in Fig.~\ref{fig:additional_probes}, which notably reaches up to $51\%$ for DeSTA2.5-Audio. GPT models show a different profile. In single-turn, GPT-4o-mini and GPT-4o
have lower reliable integration ($38\%$ and $37\%$) and larger perception
bottlenecks than the Gemini models. Their failures are therefore not primarily
Potemkin failures; they often fail earlier in the pipeline, before the
integration question becomes meaningful.

On the positional task, the reliable mass is substantially lower across all models, with even the strongest Gemini-3-Pro only reaching $31\%$. On the other hand, shortcuts and full-stack failures are much more prevalent, indicating that the judges are unable to pass the simpler perception and oracle response-mapping probes in this setting. These results suggest that current ALM judges do not yet possess the constituent perception and response-mapping abilities for temporal-causal paralinguistic judgment.

Across both single-turn and positional tasks, the judges also exhibit varying reliance on prompt cues to solve the task, which is quantified in Fig.~\ref{fig:additional_probes} as the difference in accuracy between the hard cue (single-turn) or transition cue (positional)  and the no-cue setting.


\begin{figure}[h]
    \centering
    \includegraphics[width=0.47\textwidth]{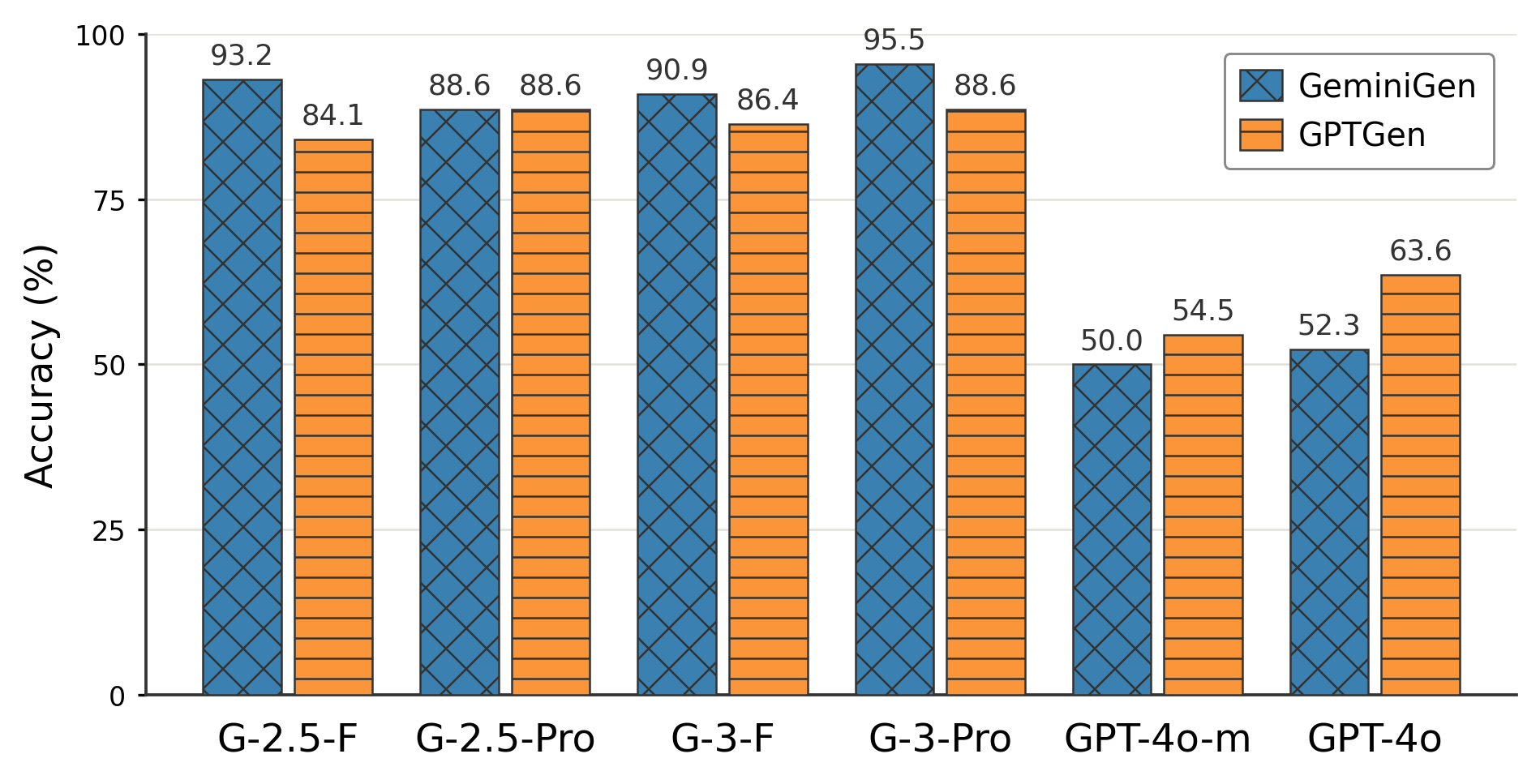} 
    \vspace{-8pt}
    \caption{\textbf{Generator bias.} Accuracy on  \textsc{Positional-emotion} comparing GeminiGen and GPTGen generation sources. The superior performance of Gemini models is due to their genuine capability, and not a bias due to involvement in the construction pipeline.}
    \label{fig:generator-judge-align}
\end{figure}

\subsection{Validity and robustness checks}
\label{sec:validity_ablation_main}
The preceding results could potentially be attributed to simpler experimental artifacts: distribution shift introduced by synthetic speech, generator-family bias, or irrelevant lexical and audio quality confounders. To ensure the findings reflect genuine capability deficits rather than methodological artifacts, we conducted a suite of targeted validity checks. We summarize these checks and provide more extensive results in the Appendix.

\begin{figure}[t]
    \centering
    \includegraphics[width=0.47\textwidth]{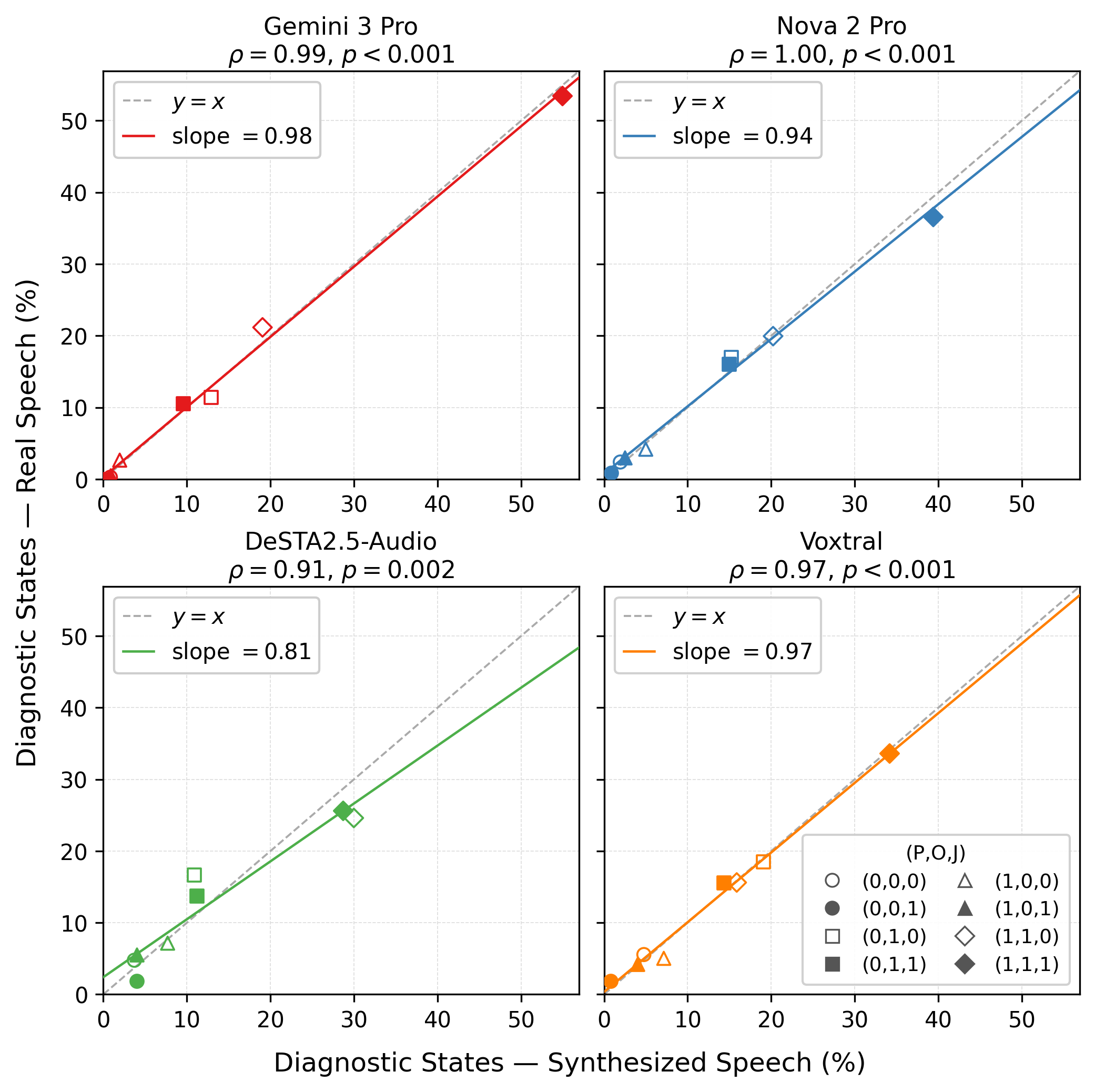} 
    \vspace{-8pt}
    \caption{\textbf{Generalization to real speech.} Comparison of $\pi_{poj}$ state proportions for synthetic vs real human speech from CAVA. The relationship has near perfect Spearman correlation and a slope close to 1 across all models.}
    \label{fig:app_single_synthetic_ablation}
\end{figure}

\noindent \textbf{Synthetic Audio Artifacts.} Replacing synthesized user audio with the original CAVA human speech yields highly correlated diagnostic profiles (Spearman $\rho>0.9$ across all models; Fig.~\ref{fig:app_single_synthetic_ablation}). This demonstrates that the observed state distribution results are not artifacts of synthetic speech. \\
\textbf{Generator Bias.} Substituting Gemini-generated positional annotations with GPT-generated annotations does not alter the qualitative model rankings, confirming that generator-judge alignment is not the primary driver of the results (Fig.~\ref{fig:generator-judge-align}).\\
\noindent \textbf{Response Quality Balance.} Differences across irrelevant audio quality or lexical factors in the response choices (e.g., one response being more polite) could mislead a judge to consistently prefer a response and perform poorly on the task. We rule out this possibility using an LLM judge (\texttt{claude-haiku-4-5}) and the DNSMOS audio quality model \cite{reddy2021dnsmosnonintrusiveperceptualobjective}, confirming that irrelevant lexical and audio qualities are balanced between candidate response choices (Tab.~\ref{tab:response-balance-check}).

%% file: sec/conclusion.tex
\section{Conclusion} \label{sec:conclusions}
We introduced an evaluation framework based on counterfactual audits to assess the capabilities of ALM judges for paralinguistic reasoning. Our experiments yield three key findings. First, contrastive recoverability overstates native judge reliability, since models can distinguish counterfactual alternatives when both are visible, but fail in the standard \textsc{Pairwise} evaluation format.
Second, aggregate accuracy hides different failure modes: the $(P,O,J)$ state distribution
separates perception bottlenecks, response-mapping failures, Potemkin failures,
and shortcut successes. Third, positional multi-turn judging is the most brittle
setting, as judges struggle when the better assistant response depends on when and why the user's affect changed. Taken together, these findings highlight the need for more extensive behavioral audits before current ALM judges are used for automatic speech-to-speech evaluation.



\section*{Limitations}

The goal of our audit is to build an evaluation framework to uncover failure modes of paralinguistic reasoning in audio judges, and we leave perspectives on improving judge performance (e.g., model training and training data design) to future work. Though we include promising experiments with real human speech, our audit relies primarily on synthesized speech. This gives the counterfactual control needed to isolate paralinguistic evidence, but it limits external validity.  Future work should test whether the same diagnostic states persist across diverse human-spoken audio, speakers, recording conditions, languages, and emotion taxonomies.

We would also like to briefly discuss some societal risks inherent to the topic of this paper. Our work could be used to improve speech-to-speech judges, which in turn could be used to improve speech-to-speech models, particularly along paralinguistic dimensions. Paralinguistically-aware voice assistants have many positive use cases, but they can also be used for harm, especially in any setting that involves emotional manipulation. We urge governments and regulatory bodies to be proactive and address these risks sooner rather than later.

%% file: sec/acknowledgments.tex
\section{Acknowledgments}
\label{sec:acknowledgments}

This research was supported by the Hariri Focused Research Program (\#2026-07-002), National Science Foundation grant CPS-2317079, and an Amazon Gift Award. We would like to thank our human study participants for their invaluable work, and Constantinos Papayiannis and Venkatesh Ravichandran for their insightful discussions.

%% file: sec/appendix.tex
\input{sec/related_work}






\section{Dataset Construction}
\label{sec:app_generation_pipeline}

\subsection{Single-Turn Dataset Construction}
\label{app:single_turn_benchmark_construction}

We use this section to go over the details of constructing the \texttt{single-turn-emotions} dataset. 

\subsubsection{Source Dataset}

In order to ground our dataset in realistic user input data, we use the Tone Awareness (EmoCF) dataset from CAVA \cite{held2025cava}. The original task provides data where each example contains  emotionally ambiguous sentences voiced by human actors in two contrasting emotions from a standard emotion taxonomy (anger, disgust, fear, joy, sadness, surprise, neutral) \cite{ravdess}. While the lexical content of the dataset is strong, we find that the human actors often fail to provide unambiguous  emotional voicings. Therefore, we opt to use TTS to synthesize new audio recordings for each example. 

\subsubsection{Filtering \& Synthesis}
We start by filtering the CAVA Tone Awareness dataset to only include obviously contrasting emotion pairs. We then synthesize each pair of user inputs with \textit{gpt-4o-mini-tts}. To generate the two voice assistant responses, we provide the following prompt to \texttt{Gemini-2.5-Flash}:
\begin{quote}\small\ttfamily
You are a friendly voice assistant.
You will be given two versions of the same sentence spoken with different emotions.
Sentence: "\{sentence\}"

Emotion A: \{emoA\}
Emotion B: \{emoB\}

For each emotion, write what a voice assistant should say in response.
The two responses should reflect the emotion they are responding to,
so that someone reading them could tell which emotion they correspond to.
However, they should not give away or directly state the emotion.
Instead, the response should strongly but implicitly acknowledge the emotion,
and it should be very clear which response corresponds to which emotion.

Respond in strict JSON format with keys "A\_response" and "B\_response".
Do not use a JSON code block. Just format your response as a JSON string.
\end{quote}

We then synthesize the two generated voice assistant responses using \texttt{gemini-2.5-flash-preview-tts}. We finally perform a manual quality check of the generated examples, resulting in the final dataset composition: 
\begin{itemize}[noitemsep, topsep=0pt,leftmargin=*]
    \item 52 samples (27.5\%) with anger vs. sadness
    \item 52 samples (27.5\%) with surprise vs. sadness
    \item 49 samples (25.9\%) with surprise vs. neutral
    \item 36 samples (19.1\%) with anger vs. surprise
\end{itemize}

\subsection{Multi-Turn Dataset Construction}
\label{app:multi_turn_benchmark_construction}

We use this section to go over the details of generating the \texttt{positional-emotion} and \texttt{emotional-conversations} multi-turn benchmarks. Both benchmarks use the same pipeline, except that the LLM prompts for transcript generation differ. 

\subsubsection{Source Conversations \& Sampling}
\label{app:source_conversations_and_sampling}

We start by sampling a source conversation $S$ from OD3 \cite{chan2023domain}, which in turn comprises text-only conversation datasets KVRET \cite{kvret}, Multi-Woz \cite{multiwoz}, DSTC11 (Track 5) \cite{zhao2023dstc11}, NOESIS-II \cite{noesisii} and SIMMC-2.1 \cite{kottur-etal-2021-simmc}. These datasets all consist of text-to-text conversations between two humans, one of whom is playing the part of a human user and the other an ``AI'' assistant that has access to a provided knowledge base. The tasks involved include finding and booking hotels, trains, restaurants, etc., shopping for clothes and furniture, getting driving directions, scheduling appointments and reminders, and planning out university coursework. OD3 adds ``repeat'' and ``rephrase'' turns to the transcripts of these conversations to make them more realistic for a voice-based conversation, although they do not explore the role of paralinguistics at all.

We sample from the training split of OD3, which has 47702 total examples, reweighting the constituent sets to bring the sampling closer to uniform. The final composition of the dataset is influenced not only by this sampling but also by which examples are rejected by the transcript-generation and audio-rendering parts of the pipeline. The final composition of the \texttt{positional-emotion} dataset is as follows (500 conversation pairs total):
\begin{itemize}[noitemsep, topsep=0pt,leftmargin=*]
    \item 171 pairs (34.2\%) from SIMMC-2.1
    \item 141 pairs (28.2\%) from NOESIS-II
    \item 112 pairs (22.4\%) from DSTC11 (Track 5)
    \item 24 pairs (4.8\%) from Multi-Woz
    \item 52 pairs (10.4\%) from KVRET
\end{itemize}

Each conversation pair yields 2 pairwise instances, giving us a total of 1000 pairwise instances in the \texttt{positional-emotion} dataset.

We note that DSTC11 (Track 5) and Multi-Woz are very similar in nature, both involving things like finding and booking restaurants, trains, hotels, etc., so together they should be viewed as a 27.2\% contingent in our benchmark. KVRET is underrepresented in \texttt{positional-emotion} because it only makes up 4.8\% of the original OD3 dataset, and many of its source conversations are too short to contain multiple failed goals.

\subsubsection{Conversation Transcript Generation Prompts}
\label{app:conversation_transcript_generation_prompts}

\input{listings/positional_emotions_generation_prompt}
\input{listings/emotional_conversations_generation_initial_prompt}
\input{listings/emotional_conversations_generation_refinement_prompt}

For the \texttt{positional-emotion} dataset, we use the LLM prompt in Lst.~\ref{lst:prompt_posemo_gen} to attempt to generate annotated text transcripts for $C_A, C_B, R_A, R_B$ from source conversation $S$. \texttt{X\_options} and \texttt{Y\_options} are sampled from the (non-empty) power sets of $\{\texttt{neutral}, \texttt{happy}\}$ and $\{\texttt{sad}, \texttt{angry}, \texttt{frazzled}, \texttt{hesitant}\}$, respectively. The voice assistant is allowed to use \texttt{neutral}, \texttt{happy}, and \texttt{empathetic} as emotions.

We also show the initial and refinement prompts for the \texttt{emotional-conversations} dataset in Lsts. \ref{lst:prompt_emoconv_init_gen} and \ref{lst:prompt_emoconv_refine_gen}, respectively. In these conversations, the human is allowed to use the emotions $\{\texttt{neutral}, \texttt{happy}, \texttt{sad}, \texttt{angry}, \texttt{hesitant}$, $\texttt{impatient}, \texttt{frazzled}\}$, and the voice assistant is allowed \texttt{neutral}, \texttt{happy}, and \texttt{empathetic}.

\subsubsection{Audio Rendering \& Quality Control}
\label{app:audio_rendering_and_qc}

\begin{table*}[t]
\centering
\begin{tabular}{|l|l|}
\hline
\textbf{Emotion} & \textbf{Audio Scorer Heuristics} \\
\hline
happy      & Use emotion2vec ``happy'' score. \\ \hline
neutral    & Use emotion2vec ``neutral'' and ``happy'' scores. \\ \hline
sad        & Use emotion2vec ``sad'' score. \\ \hline
angry      & Use emotion2vec ``angry'' and ``disgusted'' scores. \\ \hline
empathetic & Use emotion2vec ``angry'' and ``disgusted'' scores and speech rate (lower is better). \\ \hline
hesitant   & Use pause lengths and upward pitch trajectory at end. \\ \hline
impatient  & Use speech rate (higher is better) and pitch (normal is better). \\ \hline
frazzled   & Use speech rate and pitch (higher is better). \\ \hline
\end{tabular}
\caption{Heuristics used by Audio Scorer in multi-turn benchmark construction.}
\label{tab:appendix_audio_scorer_heuristics}
\end{table*}

Once we have generated annotated transcripts for $C_A, C_B, R_A, R_B$, we need to render each turn into audio to feed to the judges. To render a turn, we use a pipeline with two components, an Audio Generator and an Audio Scorer.

The Audio Generator is an OpenAI TTS (\texttt{gpt-4o-mini-tts-2025-03-20}) which is prompted with a text transcript and a natural-language description of the emotion to be rendered. For long utterances, the transcript is inserted into a ``template'' that contains further emotional cues to ensure a clear emotion. We use the following set of TTS voices: $\{\texttt{echo}, \texttt{alloy}, \texttt{ash}\}$. For each conversation, we sample one of these voices to be the human and another to be the voice assistant.

Even with very detailed emotion prompts and templates, the TTS does not always render emotions sufficiently. Thus, we use an Audio Scorer module to do automatic quality control on its outputs. The Audio Scorer gives each audio an accept/reject decision and a fitness score. These are based on scores from Speech Emotion Recognition (SER) model emotion2vec \cite{ma2023emotion2vec} as well as audio primitives such as word rate, pause length, pitch, and pitch trajectory. Our heuristics are described in Tab.~\ref{tab:appendix_audio_scorer_heuristics}, and full code for this module, along with every other part of our benchmark construction pipeline, will be released.

We generate multiple attempts for each turn, returning the highest scoring one when there have been either (a.) 3 accepted attempts, or (b.) 10 attempts total. If no attempts have been accepted at this point, we discard the turn and the entire conversation along with it.

As a final check on quality, we conduct a human validation of our benchmarks, described in Sec.~\ref{subsec:human_validation_main} in the main paper, to ensure that both the transcripts and the rendered audios are functionally sufficient for judgment.

\section{Experiment Details}
\label{sec:app_experiment_details}

\subsection{Judge Prompts}
\label{subsec:app_judge_prompts}

\subsubsection{Single-Turn Judge Prompts}

Here we provide the judge prompts used for each experiment in the \texttt{single-turn-emotions} benchmark. Since there are minimal differences between the matching and pairwise formats, we only provide the pairwise prompts here. Additionally, each prompt can take one of the following cues:
\begin{enumerate}[noitemsep, topsep=0pt, leftmargin=*]
    \item \textbf{No Cue:} {\small\ttfamily \textbackslash n} (i.e., no additional instruction)
    \item \textbf{Soft Cue:} {\small\ttfamily Focus on tone of voice when making your decision.\textbackslash n}
    \item \textbf{Hard Cue:} {\small\ttfamily Focus *only* on tone of voice, and not words, when making your decision.\textbackslash n}
\end{enumerate}

For the main performance results reported in Table~\ref{tab:headline-results}, we use the following prompt: 

\begin{quote}\small\ttfamily
You will hear audio clip C, and then be given two possible voice assistant replies, R1 and R2. 
Your task: decide which response is more appropriate for the conversation C based on user experience.

\{cue\}

Reply with a JSON dict with entries "reasoning", containing your reasoning, and "pred", containing your prediction, 
which should be either "R1" or "R2".
Do not include code blocks for the JSON. Respond with the JSON string only.

C:
\{C.wav\}

R1:
\{R1.wav\}

R2:
\{R2.wav\}
\end{quote}

We then test the model's ability to explicitly predict the emotion in the user's audio (Table~\ref{tab:full_single_explicit}), which we term the \textbf{Perception} task: 
\begin{quote}\small\ttfamily
You will hear audio clip C, and then be given the names of two emotions, R1 and R2. Your task: decide which emotion more accurately describes audio clip C.

\{cue\}

Reply with a JSON dict with entries "reasoning", containing your reasoning, and "pred", containing your prediction, 
which should be either "R1" or "R2".
Do not include code blocks for the JSON. Respond with the JSON string only.

R1 = \{emotion\_1\} \\
R2 = \{emotion\_2\}

C:
\{C.wav\}
\end{quote}

Next, we experiment with the model's ability to judge the correct response given the ground truth emotion label. This is termed the \textbf{Reasoning} task, which is also called the \textbf{Oracle} orchestration approach in the main text. The results are reported in Table~\ref{tab:full_single_oracle}. Crucially, this experiment is conducted using text alone: 

\begin{quote}\small\ttfamily
You will be given a user input C, and then be given two possible voice assistant replies, R1 and R2. 
The user input will have a text transcript ("transcript"), and a description of the speaker's emotional tone of voice ("emotion"). 
Your task: select which response is more appropriate for the conversation based on user experience.

\{cue\}

Reply with a JSON dict with entries "reasoning", containing your reasoning, and "pred", containing your prediction, 
which should be either "R1" or "R2".
Do not include code blocks for the JSON. Respond with the JSON string only.

C:
  transcript: \{transcript\}
  emotion: \{emotion\}

R1: \{r1\_response\}

R2: \{r2\_response\}
\end{quote}

Finally, we experiment with the Staged approach reported in Table~\ref{tab:full_single_staging}. Crucially, this experiment is also conducted using text alone: 

\begin{quote}\small\ttfamily
You will be given a user input C, and then be given two possible voice assistant replies, R1 and R2. 
The user input will have a text transcript ("transcript"), and a description of the speaker's emotional tone of voice ("emotion"). 
Keep in mind that the human's emotions were detected by an audio-language model which does not have perfect accuracy.

Your task: select which response is more appropriate for the conversation based on user experience.

\{cue\}

Reply with a JSON dict with entries "reasoning", containing your reasoning, and "pred", containing your prediction, 
which should be either "R1" or "R2".
Do not include code blocks for the JSON. Respond with the JSON string only.

C:
  transcript: \{transcript\}
  emotion: \{emo\_pred\}

R1: \{r1\_response\}

R2: \{r2\_response\}
\end{quote}

\subsubsection{Multi-Turn Judge Prompts}

We show the code that produces the judge prompts for multi-turn experiments. The end-to-end judgment (J) prompts are shown in Lst.~\ref{lst:make_prompt}. The perception (P) probe prompts are shown in Lst.~\ref{lst:make_prompt_multi_emotion}. The oracle (O) and Staging probe prompts are shown in Lst.~\ref{lst:make_prompt_for_LLM}.

\input{listings/make_prompt}
\input{listings/make_multi_emotion_prompt}
\input{listings/make_prompt_for_LLM}

\section{Extended Results}
\label{sec:app_extended_results}

\subsection{Confidence Intervals and Protocol Gap Analysis}
\label{subsec:app_CIs}

\paragraph{Main results with confidence intervals:} We report the results from Tab.~\ref{tab:headline-results} with 95\% Wilson confidence intervals in Tab.~\ref{tab:ci_main_single} for \texttt{single-turn-emotions} and Tab.~\ref{tab:ci_main_positional} for \texttt{positional-emotion}.

\paragraph{Protocol gap analysis:} We quantify protocol collapse by defining a metric, \(\Delta_{\mathrm{protocol}}\), which is the Matching accuracy minus the Pairwise accuracy for a given judge and prompt. We report \(\Delta_{\mathrm{protocol}}\) with paired bootstrap intervals over underlying counterfactual items, resampling items with replacement and recomputing \(\Delta_{\mathrm{protocol}}\) on each bootstrap replicate. Results of this analysis are in Tab.~\ref{tab:protocol-gap-ci}. We see that the Gemini family models experience significant protocol collapse for both the ``No'' and ``Hard'' paralinguistic prompt cues on the \texttt{single-turn-emotions} dataset and for the ``Trans'' prompt cue on the \texttt{positional-emotion} dataset. Under these settings, Gemini judges can extract and use differential signals when given both audio contexts but collapse to random-chance performance when only given one context, as they would in a deployment setting. Judges other than Gemini have a small gap, but only because they get random-chance performance \textit{even when given both audio contexts}, so there is nothing left to collapse. A similar phenomenon happens for Gemini family models on \texttt{positional-emotion} for the ``No'' prompt cue. This means that Gemini judges need to be told explicitly to look for an emotion transition, \textit{and} given counterfactual inputs (which would not be available in deployment), to get above-random performance on \texttt{positional-emotion}.
\begin{table*}[h]
\centering\small\setlength{\tabcolsep}{4pt}
\begin{tabular}{lcccc}
\toprule
Judge & Single No & Single Hard & Positional No & Positional Trans. \\
\midrule
Gemini-2.5-Flash & 21.3 [12.9, 29.4] &  15.0 [6.7, 23.0] &  0.3 [-5.1, 5.6] &  8.1 [3.0, 13.3] \\
Gemini-2.5-Pro & 28.4 [17.9, 38.5] &  26.0 [17.0, 34.5] &  7.4 [2.2, 12.7] &  11.3 [6.1, 16.5] \\
Gemini-3-Flash & 27.0 [19.3, 34.7] &  24.3 [16.9, 31.5]  &  3.3 [-1.9, 8.6] &  26.4 [21.7, 31.1] \\
Gemini-3-Pro & 29.2 [21.9, 36.2] &  25.7 [19.3, 31.8] &  6.3 [1.0, 11.7] &  14.8 [9.7, 19.9] \\
GPT-4o-mini & -7.0 [-16.0, 1.9] &  -0.5 [-9.4, 8.3] &  -3.7 [-9.0, 1.7] &  0.0 [-5.4, 5.5] \\
GPT-4o & 2.8 [-6.1, 11.7] &  2.0 [-6.8, 10.5] &  -3.6 [-9.0, 1.7] &  2.4 [-2.9, 7.7] \\
Nova-2-Pro & 6.7 [-2.8, 16.2] &  -4.8 [-13.4, 3.9]  &  -1.1 [-6.5, 4.3] &  2.2 [-3.0, 7.6] \\
Qwen2.5-Omni-7B &  1.7 [-7.0, 10.4] &  7.0 [-1.9, 15.8] &  -0.2 [-5.5, 5.2] &  -2.7 [-8.2, 2.7] \\
Voxtral-Small-24B &  1.0 [-8.7, 10.9] &  0.8 [-7.9, 9.3] &  -1.2 [-6.8, 4.3] &  -2.5 [-7.9, 2.8] \\
DeSTA2.5-Audio & 0.2 [-8.6, 9.1] &  -1.8 [-11.5, 7.9]  &  --  &  --  \\
Phi-4-Multimodal-6B &  -4.4 [-13.2, 4.4] &  0.5 [-8.5, 9.5] &  -6.4 [-12.2, -0.7] &  -2.4 [-7.9, 3.1] \\
\bottomrule
\end{tabular}
\caption{\textbf{Protocol gap}: Matching accuracy minus Pairwise accuracy, with paired bootstrap 95\% confidence intervals over underlying counterfactual items. Positive values mean the judge succeeds when both counterfactual audios/responses are visible but degrades in the one-context Pairwise setting.}
\label{tab:protocol-gap-ci}
\end{table*}

\begin{table*}[h]
\centering\small\setlength{\tabcolsep}{4pt}
\begin{tabular}{lcccc}
\toprule
Judge & Protocol & Cue & n & Accuracy [95\% CI] \\
\midrule
Gemini-2.5-Flash & Pairwise & No & 375 & 53.1 [48.0, 58.1] \\
Gemini-2.5-Flash & Pairwise & Hard & 378 & 59.0 [54.0, 63.8] \\
Gemini-2.5-Flash & Matching & No & 183 & 74.3 [67.5, 80.1] \\
Gemini-2.5-Flash & Matching & Hard & 173 & 74.0 [67.0, 80.0] \\
Gemini-2.5-Pro & Pairwise & No & 211 & 52.1 [45.4, 58.8] \\
Gemini-2.5-Pro & Pairwise & Hard & 353 & 60.1 [54.9, 65.0] \\
Gemini-2.5-Pro & Matching & No & 103 & 80.6 [71.9, 87.1] \\
Gemini-2.5-Pro & Matching & Hard & 93 & 86.0 [77.5, 91.6] \\
Gemini-3-Flash & Pairwise & No & 378 & 52.4 [47.3, 57.4] \\
Gemini-3-Flash & Pairwise & Hard & 378 & 58.7 [53.7, 63.6] \\
Gemini-3-Flash & Matching & No & 189 & 79.4 [73.0, 84.5] \\
Gemini-3-Flash & Matching & Hard & 189 & 83.1 [77.1, 87.7] \\
Gemini-3-Pro & Pairwise & No & 377 & 55.2 [50.1, 60.1] \\
Gemini-3-Pro & Pairwise & Hard & 377 & 65.3 [60.3, 69.9] \\
Gemini-3-Pro & Matching & No & 185 & 84.3 [78.4, 88.9] \\
Gemini-3-Pro & Matching & Hard & 188 & 91.0 [86.0, 94.3] \\
GPT-4o-mini & Pairwise & No & 377 & 52.3 [47.2, 57.2] \\
GPT-4o-mini & Pairwise & Hard & 377 & 55.4 [50.4, 60.4] \\
GPT-4o-mini & Matching & No & 179 & 45.3 [38.1, 52.6] \\
GPT-4o-mini & Matching & Hard & 184 & 54.9 [47.7, 61.9] \\
GPT-4o & Pairwise & No & 378 & 54.2 [49.2, 59.2] \\
GPT-4o & Pairwise & Hard & 377 & 53.8 [48.8, 58.8] \\
GPT-4o & Matching & No & 170 & 57.1 [49.5, 64.3] \\
GPT-4o & Matching & Hard & 188 & 55.9 [48.7, 62.8] \\
Nova-2-Pro & Pairwise & No & 378 & 48.4 [43.4, 53.4] \\
Nova-2-Pro & Pairwise & Hard & 377 & 58.4 [53.3, 63.2] \\
Nova-2-Pro & Matching & No & 147 & 55.1 [47.0, 62.9] \\
Nova-2-Pro & Matching & Hard & 183 & 53.6 [46.3, 60.6] \\
Qwen2.5-Omni-7B & Pairwise & No & 378 & 50.0 [45.0, 55.0] \\
Qwen2.5-Omni-7B & Pairwise & Hard & 378 & 47.1 [42.1, 52.1] \\
Qwen2.5-Omni-7B & Matching & No & 180 & 51.7 [44.4, 58.9] \\
Qwen2.5-Omni-7B & Matching & Hard & 185 & 54.1 [46.9, 61.1] \\
Voxtral-Small-24B & Pairwise & No & 378 & 52.9 [47.9, 57.9] \\
Voxtral-Small-24B & Pairwise & Hard & 378 & 53.2 [48.1, 58.1] \\
Voxtral-Small-24B & Matching & No & 128 & 53.9 [45.3, 62.3] \\
Voxtral-Small-24B & Matching & Hard & 189 & 54.0 [46.9, 60.9] \\
DeSTA2.5-Audio & Pairwise & No & 378 & 48.9 [43.9, 54.0] \\
DeSTA2.5-Audio & Pairwise & Hard & 378 & 47.9 [42.9, 52.9] \\
DeSTA2.5-Audio & Matching & No & 185 & 49.2 [42.1, 56.3] \\
DeSTA2.5-Audio & Matching & Hard & 141 & 46.1 [38.1, 54.3] \\
Phi-4-Multimodal-6B & Pairwise & No & 377 & 53.6 [48.5, 58.6] \\
Phi-4-Multimodal-6B & Pairwise & Hard & 378 & 48.7 [43.7, 53.7] \\
Phi-4-Multimodal-6B & Matching & No & 187 & 49.2 [42.1, 56.3] \\
Phi-4-Multimodal-6B & Matching & Hard & 177 & 49.2 [41.9, 56.5] \\
\bottomrule
\end{tabular}
\caption{Single-turn end-to-end accuracy with Wilson 95\% confidence intervals.}
\label{tab:ci_main_single}
\end{table*}

\begin{table*}[h]
\centering\small\setlength{\tabcolsep}{4pt}
\begin{tabular}{lcccc}
\toprule
Judge & Protocol & Cue & n & Accuracy [95\% CI] \\
\midrule
Gemini-2.5-Flash & Pairwise & No & 1000 & 48.7 [45.6, 51.8] \\
Gemini-2.5-Flash & Pairwise & Trans & 1000 & 53.9 [50.8, 57.0] \\
Gemini-2.5-Flash & Matching & No & 498 & 49.0 [44.6, 53.4] \\
Gemini-2.5-Flash & Matching & Trans & 500 & 62.0 [57.7, 66.1] \\
Gemini-2.5-Pro & Pairwise & No & 1000 & 50.2 [47.1, 53.3] \\
Gemini-2.5-Pro & Pairwise & Trans & 1000 & 54.5 [51.4, 57.6] \\
Gemini-2.5-Pro & Matching & No & 498 & 57.6 [53.2, 61.9] \\
Gemini-2.5-Pro & Matching & Trans & 500 & 65.8 [61.5, 69.8] \\
Gemini-3-Flash & Pairwise & No & 1000 & 50.3 [47.2, 53.4] \\
Gemini-3-Flash & Pairwise & Trans & 1000 & 53.0 [49.9, 56.1] \\
Gemini-3-Flash & Matching & No & 500 & 53.6 [49.2, 57.9] \\
Gemini-3-Flash & Matching & Trans & 500 & 79.4 [75.6, 82.7] \\
Gemini-3-Pro & Pairwise & No & 1000 & 48.7 [45.6, 51.8] \\
Gemini-3-Pro & Pairwise & Trans & 1000 & 51.6 [48.5, 54.7] \\
Gemini-3-Pro & Matching & No & 500 & 55.0 [50.6, 59.3] \\
Gemini-3-Pro & Matching & Trans & 500 & 66.4 [62.1, 70.4] \\
GPT-4o-mini & Pairwise & No & 1000 & 52.9 [49.8, 56.0] \\
GPT-4o-mini & Pairwise & Trans & 1000 & 52.0 [48.9, 55.1] \\
GPT-4o-mini & Matching & No & 500 & 49.2 [44.8, 53.6] \\
GPT-4o-mini & Matching & Trans & 500 & 52.0 [47.6, 56.3] \\
GPT-4o & Pairwise & No & 1000 & 50.3 [47.2, 53.4] \\
GPT-4o & Pairwise & Trans & 1000 & 49.2 [46.1, 52.3] \\
GPT-4o & Matching & No & 499 & 46.7 [42.4, 51.1] \\
GPT-4o & Matching & Trans & 500 & 51.6 [47.2, 56.0] \\
Nova-2-Pro & Pairwise & No & 982 & 50.7 [47.6, 53.8] \\
Nova-2-Pro & Pairwise & Trans & 990 & 48.3 [45.2, 51.4] \\
Nova-2-Pro & Matching & No & 498 & 49.6 [45.2, 54.0] \\
Nova-2-Pro & Matching & Trans & 499 & 50.5 [46.1, 54.9] \\
Qwen2.5-Omni-7B & Pairwise & No & 1000 & 47.3 [44.2, 50.4] \\
Qwen2.5-Omni-7B & Pairwise & Trans & 996 & 51.1 [48.0, 54.2] \\
Qwen2.5-Omni-7B & Matching & No & 499 & 47.1 [42.8, 51.5] \\
Qwen2.5-Omni-7B & Matching & Trans & 490 & 48.4 [44.0, 52.8] \\
Voxtral-Small-24B & Pairwise & No & 990 & 49.7 [46.6, 52.8] \\
Voxtral-Small-24B & Pairwise & Trans & 994 & 49.7 [46.6, 52.8] \\
Voxtral-Small-24B & Matching & No & 466 & 48.5 [44.0, 53.0] \\
Voxtral-Small-24B & Matching & Trans & 500 & 47.2 [42.9, 51.6] \\
DeSTA2.5-Audio & Pairwise & No & 810 & 50.6 [47.2, 54.0] \\
DeSTA2.5-Audio & Pairwise & Trans & 958 & 50.3 [47.2, 53.5] \\
Phi-4-Multimodal-6B & Pairwise & No & 864 & 51.6 [48.3, 54.9] \\
Phi-4-Multimodal-6B & Pairwise & Trans & 922 & 49.3 [46.1, 52.6] \\
Phi-4-Multimodal-6B & Matching & No & 456 & 45.2 [40.7, 49.8] \\
Phi-4-Multimodal-6B & Matching & Trans & 481 & 47.0 [42.6, 51.5] \\
\bottomrule
\end{tabular}
\caption{\texttt{positional-emotion} end-to-end accuracy with Wilson 95\% confidence intervals.}
\label{tab:ci_main_positional}
\end{table*}

\subsection{Full Extended Results}
\label{subsec:app_full_extended_results}

We use this section to show full extended results of our experiments. These include an extended set of judge prompt cues (see Sec.~\ref{subsec:app_judge_prompts} for full description of all prompts and cues), as well as an additional \textbf{Staging} probe in which we feed the judge its own predicted emotions from the Perception (P) probe. We also provide baseline results for the \texttt{emotional-conversations} dataset. 

For the \texttt{single-turn-emotions} dataset, we have end-to-end Pairwise accuracies in Tab.~\ref{tab:full_single_pairwise}, Matching accuracies in Tab.~\ref{tab:full_single_matching}, Oracle (O) probe accuracies in Tab.~\ref{tab:full_single_oracle}, Perception (P) probe per-turn accuracies in Tab.~\ref{tab:full_single_explicit}, and Staging probe accuracies in Tab.~\ref{tab:full_single_staging}.

For the \texttt{emotional-conversations} dataset, we show end-to-end accuracies in the matching and pairwise settings in Tab.~\ref{tab:appendix_emotional_conversations_matching} and \ref{tab:appendix_emotional_conversations_pairwise} respectively. We find that the last turn alone is sufficient to solve this task, indicating that current ALMs are robust to multi-turn history when the relevant paralinguistic state is only in the last turn. 

For the \texttt{positional-emotion} dataset, we have end-to-end pairwise accuracies in Tab.~\ref{tab:full_positional_pairwise}, Matching accuracies in Tab.~\ref{tab:full_positional_matching}, Oracle (O) probe accuracies in Tab.~\ref{tab:full_positional_oracle}, Perception (P) probe per-turn accuracies in Tab.~\ref{tab:full_positional_explicit}, and Staging probe accuracies in Tab.~\ref{tab:full_positional_staging}.

\begin{table*}[t]
\centering
\begin{tabular}{l|ccc|ccc}
\toprule
& \multicolumn{3}{c|}{Full Conversation}
& \multicolumn{3}{c}{Last Turn Only} \\
Judge
& No Cue & Soft Cue & Hard Cue
& No Cue & Soft Cue & Hard Cue \\
\midrule
Gemini-2.5-Flash & 68.4 & 74.8 & 75.2 & 68.0 & 76.4 & 75.6 \\
Gemini-2.5-Pro   & 83.6 & 91.2 & 92.0 & 84.4 & 86.8 & 87.2 \\
Gemini-3-Flash   & 82.4 & 92.8 & 94.8 & 86.8 & 89.6 & 90.8 \\
Gemini-3-Pro     & 89.6 & 97.2 & 96.4 & 95.2 & 96.4 & 96.4 \\
GPT-4o-mini      & 49.6 & 52.0 & 54.0 & 51.6 & 49.2 & 52.0 \\
GPT-4o           & 51.6 & 52.4 & 48.4 & 49.2 & 56.8 & 55.6 \\
\bottomrule
\end{tabular}
\caption{\textbf{Matching end-to-end judgment (J)} accuracy for \texttt{emotional-conversations}.}
\label{tab:appendix_emotional_conversations_matching}
\end{table*}

\begin{table*}[t]
\centering
\begin{tabular}{l|ccc|ccc}
\toprule
& \multicolumn{3}{c|}{Full Conversation}
& \multicolumn{3}{c}{Last Turn Only} \\
Judge
& No Cue & Soft Cue & Hard Cue
& No Cue & Soft Cue & Hard Cue \\
\midrule
Gemini-2.5-Flash & 54.4 & 59.8 & 62.6 & 53.0 & 57.6 & 62.0 \\
Gemini-2.5-Pro   & 53.6 & 66.4 & 71.0 & 54.4 & 66.2 & 71.2 \\
Gemini-3-Flash   & 59.2 & 66.2 & 70.0 & 57.8 & 64.6 & 70.0 \\
Gemini-3-Pro     & 67.6 & 74.8 & 77.6 & 65.0 & 74.2 & 75.4 \\
GPT-4o-mini      & 49.0 & 51.2 & 56.8 & 51.6 & 51.0 & 54.0 \\
GPT-4o           & 51.6 & 56.6 & 55.6 & 52.8 & 57.0 & 57.6 \\
\bottomrule
\end{tabular}
\caption{\textbf{Pairwise end-to-end judgment (J)} accuracy for \texttt{emotional-conversations}.}
\label{tab:appendix_emotional_conversations_pairwise}
\end{table*}

\begin{table*}[ht]
\centering
\begin{tabular}{lccc}
\toprule
Judge & No Cue & Soft Cue & Hard Cue \\
\midrule
Gemini-2.5-Flash & 53.1 & 54.8 & 59.0 \\
Gemini-2.5-Pro & 52.1 & 61.8 & 60.1 \\
Gemini-3-Flash & 52.4 & 59.0 & 58.7 \\
Gemini-3-Pro & 55.2 & 58.4 & 65.3 \\
GPT-4o-mini & 52.3 & 50.5 & 55.4 \\
GPT-4o & 54.2 & 53.5 & 53.8 \\
Nova-2-Pro & 48.4 & 53.3 & 58.4 \\
Qwen2.5-Omni-7B & 50.0 & 51.6 & 47.1 \\
Voxtral-Small-24B & 52.9 & 53.4 & 53.2 \\
DeSTA2.5-Audio & 48.9 & 52.5 & 47.9 \\
Phi-4-Multimodal-6B & 53.6 & 50.0 & 48.7 \\
\bottomrule
\end{tabular}
\caption{\textbf{Pairwise end-to-end judgment (J)} accuracy for \texttt{single-turn-emotions}.}
\label{tab:full_single_pairwise}
\end{table*}

\begin{table*}[ht]
\centering
\begin{tabular}{lccc}
\toprule
Judge & No Cue & Soft Cue & Hard Cue \\
\midrule
Gemini-2.5-Flash & 74.3 & 76.7 & 74.0 \\
Gemini-2.5-Pro & 80.6 & 85.3 & 86.0 \\
Gemini-3-Flash & 79.4 & 81.5 & 83.1 \\
Gemini-3-Pro & 84.3 & 90.4 & 91.0 \\
GPT-4o-mini & 45.3 & 51.6 & 54.9 \\
GPT-4o & 57.1 & 57.8 & 55.9 \\
Nova-2-Pro & 55.1 & 62.2 & 53.6 \\
Qwen2.5-Omni-7B & 51.7 & 43.6 & 54.1 \\
Voxtral-Small-24B & 53.9 & 50.3 & 54.0 \\
DeSTA2.5-Audio & 49.2 & 47.2 & 46.1 \\
Phi-4-Multimodal-6B & 49.2 & 48.6 & 49.2 \\
\bottomrule
\end{tabular}
\caption{\textbf{Matching end-to-end judgment (J)} accuracy for \texttt{single-turn-emotions}.}
\label{tab:full_single_matching}
\end{table*}

\begin{table*}[ht]
\centering
\begin{tabular}{lccc}
\toprule
Judge & No Cue & Soft Cue & Hard Cue \\
\midrule
Gemini-2.5-Flash & 78.2 & 82.0 & 95.5 \\
Gemini-2.5-Pro & 81.3 & 89.7 & 94.3 \\
Gemini-3-Flash & 83.9 & 88.4 & 96.6 \\
Gemini-3-Pro & 82.8 & 90.7 & 96.6 \\
GPT-4o-mini & 90.7 & 93.9 & 96.0 \\
GPT-4o & 83.6 & 85.3 & 92.6 \\
Nova-2-Pro & 77.7 & 84.1 & 89.7 \\
Qwen2.5-Omni-7B & 83.1 & 84.4 & 84.7 \\
Voxtral-Small-24B & 73.8 & 79.1 & 83.3 \\
DeSTA2.5-Audio & 73.3 & 81.0 & 80.7 \\
Phi-4-Multimodal-6B & 75.9 & 77.2 & 78.1 \\
\bottomrule
\end{tabular}
\caption{\textbf{Oracle (O) probe} accuracy for \texttt{single-turn-emotions}, in which the judge is given text transcript annotated with ground-truth emotions.}
\label{tab:full_single_oracle}
\end{table*}

\begin{table*}[ht]
\centering
\begin{tabular}{lccc}
\toprule
Judge & No Cue & Soft Cue & Hard Cue \\
\midrule
Gemini-2.5-Flash & 71.3 & 71.4 & 73.5 \\
Gemini-2.5-Pro & 75.3 & 78.8 & 82.0 \\
Gemini-3-Flash & 73.3 & 75.4 & 77.2 \\
Gemini-3-Pro & 74.0 & 72.4 & 76.5 \\
GPT-4o-mini & 50.5 & 57.3 & 57.7 \\
GPT-4o & 60.2 & 60.4 & 59.0 \\
Nova-2-Pro & 60.1 & 64.1 & 67.1 \\
Qwen2.5-Omni-7B & 55.7 & 59.7 & 57.4 \\
Voxtral-Small-24B & 57.4 & 63.0 & 61.1 \\
DeSTA2.5-Audio & 67.7 & 69.3 & 70.3 \\
Phi-4-Multimodal-6B & 51.7 & 52.9 & 54.1 \\
\bottomrule
\end{tabular}
\caption{\textbf{Perception (P) probe} per-turn accuracy for \texttt{single-turn-emotions}. This is the average percentage of human turns for which the judge predicts the correct emotion when explicitly asked to do so.}
\label{tab:full_single_explicit}
\end{table*}

\begin{table*}[ht]
\centering
\begin{tabular}{lccc}
\toprule
Judge & No Cue & Soft Cue & Hard Cue \\
\midrule
Gemini-2.5-Flash & 59.3 & 64.8 & 70.8 \\
Gemini-2.5-Pro & 63.1 & 71.3 & 78.4 \\
Gemini-3-Flash & 63.8 & 68.3 & 77.0 \\
Gemini-3-Pro & 62.3 & 65.1 & 74.3 \\
GPT-4o-mini & 48.8 & 56.1 & 56.2 \\
GPT-4o & 54.4 & 53.5 & 56.1 \\
Nova-2-Pro & 54.1 & 61.2 & 63.8 \\
Qwen2.5-Omni-7B & 54.1 & 54.8 & 52.6 \\
Voxtral-Small-24B & 55.0 & 58.7 & 56.9 \\
DeSTA2.5-Audio & 59.5 & 62.3 & 64.7 \\
Phi-4-Multimodal-6B & 50.5 & 53.2 & 53.8 \\
\bottomrule
\end{tabular}
\caption{\textbf{Staging probe} accuracy for \texttt{single-turn-emotions}, in which we feed the judge a text-transcript annotated with the judge's predicted emotions from Tab.~\ref{tab:full_single_explicit}}
\label{tab:full_single_staging}
\end{table*}

\begin{table*}[ht]
\centering
\begin{tabular}{lcccc}
\toprule
Judge & No Cue & Soft Cue & Hard Cue & Transition Cue \\
\midrule
Gemini-2.5-Flash & 48.7 & 49.4 & 50.0 & 53.9\\
Gemini-2.5-Pro & 50.2 & 48.8 & 53.5 & 54.5\\
Gemini-3-Flash & 50.3 & 49.2 & 51.2 & 53.0\\
Gemini-3-Pro & 48.7 & 50.6 & 51.5 & 51.6\\
GPT-4o-mini & 52.9 & 53.5 & 53.2 & 52.0\\
GPT-4o & 50.3 & 50.0 & 49.9 & 49.2\\
Nova-2-Pro & 50.7 & 52.3 & 49.2 & 48.3\\
Qwen2.5-Omni-7B & 47.3 & 48.1 & 48.3 & 51.1\\
Voxtral-Small-24B & 49.7 & 50.6 & 50.7 & 49.7\\
DeSTA2.5-Audio & 50.6 & 52.7 & 48.3 & 50.3\\
Phi-4-Multimodal-6B & 51.6 & 47.9 & 51.5 & 49.3\\
\bottomrule
\end{tabular}
\caption{\textbf{Pairwise end-to-end judgment (J)} accuracy for \texttt{positional-emotion}.}
\label{tab:full_positional_pairwise}
\end{table*}

\begin{table*}[ht]
\centering
\begin{tabular}{lcccc}
\toprule
Judge & No Cue & Soft Cue & Hard Cue & Transition Cue \\
\midrule
Gemini-2.5-Flash & 49.0 & 49.0 & 50.8 & 62.0\\
Gemini-2.5-Pro & 57.6 & 54.4 & 52.0 & 65.8\\
Gemini-3-Flash & 53.6 & 47.2 & 48.2 & 79.4\\
Gemini-3-Pro & 55.0 & 54.2 & 54.8 & 66.4\\
GPT-4o-mini & 49.2 & 51.0 & 51.4 & 52.0\\
GPT-4o & 46.7 & 52.4 & 50.0 & 51.6\\
Nova-2-Pro & 49.6 & 49.5 & 51.6 & 50.5\\
Qwen2.5-Omni-7B & 47.1 & 48.2 & 47.2 & 48.4\\
Voxtral-Small-24B & 48.5 & 49.5 & 50.6 & 47.2\\
Phi-4-Multimodal-6B & 45.2 & 46.0 & 47.5 & 47.0\\
\bottomrule
\end{tabular}
\caption{\textbf{Matching end-to-end judgment (J)} accuracy for \texttt{positional-emotion}.}
\label{tab:full_positional_matching}
\end{table*}

\begin{table*}[ht]
\centering
\begin{tabular}{lcccc}
\toprule
Judge & No Cue & Soft Cue & Hard Cue & Transition Cue \\
\midrule
Gemini-2.5-Flash & 51.2 & 51.4 & 54.5 & 67.1\\
Gemini-2.5-Pro & 54.9 & 55.8 & 62.6 & 70.1\\
Gemini-3-Flash & 54.7 & 56.1 & 59.5 & 69.2\\
Gemini-3-Pro & 53.4 & 55.9 & 61.8 & 70.3\\
GPT-4o-mini & 47.9 & 47.6 & 47.6 & 49.2\\
GPT-4o & 49.4 & 48.2 & 49.2 & 52.3\\
Nova-2-Pro & 49.6 & 46.3 & 49.2 & 52.3\\
Qwen2.5-Omni-7B & 49.6 & 48.7 & 48.6 & 49.3\\
Voxtral-Small-24B & 50.3 & 49.2 & 50.3 & 56.4\\
DeSTA2.5-Audio & 47.5 & 48.2 & 47.5 & 50.2\\
Phi-4-Multimodal-6B & 47.4 & 50.0 & 46.5 & 50.5\\
\bottomrule
\end{tabular}
\caption{\textbf{Oracle (O) probe} accuracy for \texttt{positional-emotion}, in which the judge is given text transcript annotated with ground-truth emotions.}
\label{tab:full_positional_oracle}
\end{table*}

\begin{table*}[ht]
\centering
\begin{tabular}{lcccc}
\toprule
Judge & No Cue & Soft Cue & Hard Cue & Transition Cue \\
\midrule
Gemini-2.5-Flash & 60.8 & 63.9 & 64.5 & 81.8\\
Gemini-2.5-Pro & 76.0 & 79.8 & 81.6 & 88.0\\
Gemini-3-Flash & 79.9 & 81.6 & 82.8 & 88.4\\
Gemini-3-Pro & 83.6 & 85.1 & 87.8 & 88.9\\
GPT-4o-mini & 50.2 & 51.3 & 51.1 & 54.2\\
GPT-4o & 53.4 & 53.3 & 54.1 & 59.6\\
Nova-2-Pro & 56.2 & 57.9 & 58.9 & 65.5\\
Voxtral-Small-24B & 50.8 & 51.1 & 51.1 & 59.4\\
\bottomrule
\end{tabular}
\caption{\textbf{Perception (P) probe} per-turn accuracy for \texttt{positional-emotion}. This is the average percentage of human turns for which the judge predicts the correct emotion when explicitly asked to do so.}
\label{tab:full_positional_explicit}
\end{table*}

\begin{table*}[ht]
\centering
\begin{tabular}{lcccc}
\toprule
Judge & No Cue & Soft Cue & Hard Cue & Transition Cue \\
\midrule
Gemini-2.5-Flash & 52.0 & 53.1 & 51.7 & 56.4\\
Gemini-2.5-Pro & 50.8 & 52.2 & 59.1 & 59.3\\
Gemini-3-Flash & 51.7 & 52.2 & 56.4 & 59.8\\
Gemini-3-Pro & 51.6 & 53.1 & 55.9 & 61.9\\
GPT-4o-mini & 48.0 & 51.5 & 50.6 & 47.5\\
GPT-4o & 48.6 & 48.6 & 47.5 & 49.3\\
Nova-2-Pro & 48.7 & 49.7 & 49.5 & 49.7\\
Voxtral-Small-24B & 50.5 & 51.5 & 48.4 & 49.5\\
\bottomrule
\end{tabular}
\caption{\textbf{Staging probe} accuracy for \texttt{positional-emotion}, in which we feed the judge a text-transcript annotated with the judge's predicted emotions from Tab.~\ref{tab:full_positional_explicit}}
\label{tab:full_positional_staging}
\end{table*}

\section{Human Validation Browser Interfaces}
\label{app:human_validation}

\begin{figure*}[t]
  \centering
  \includegraphics[width=0.7\linewidth]{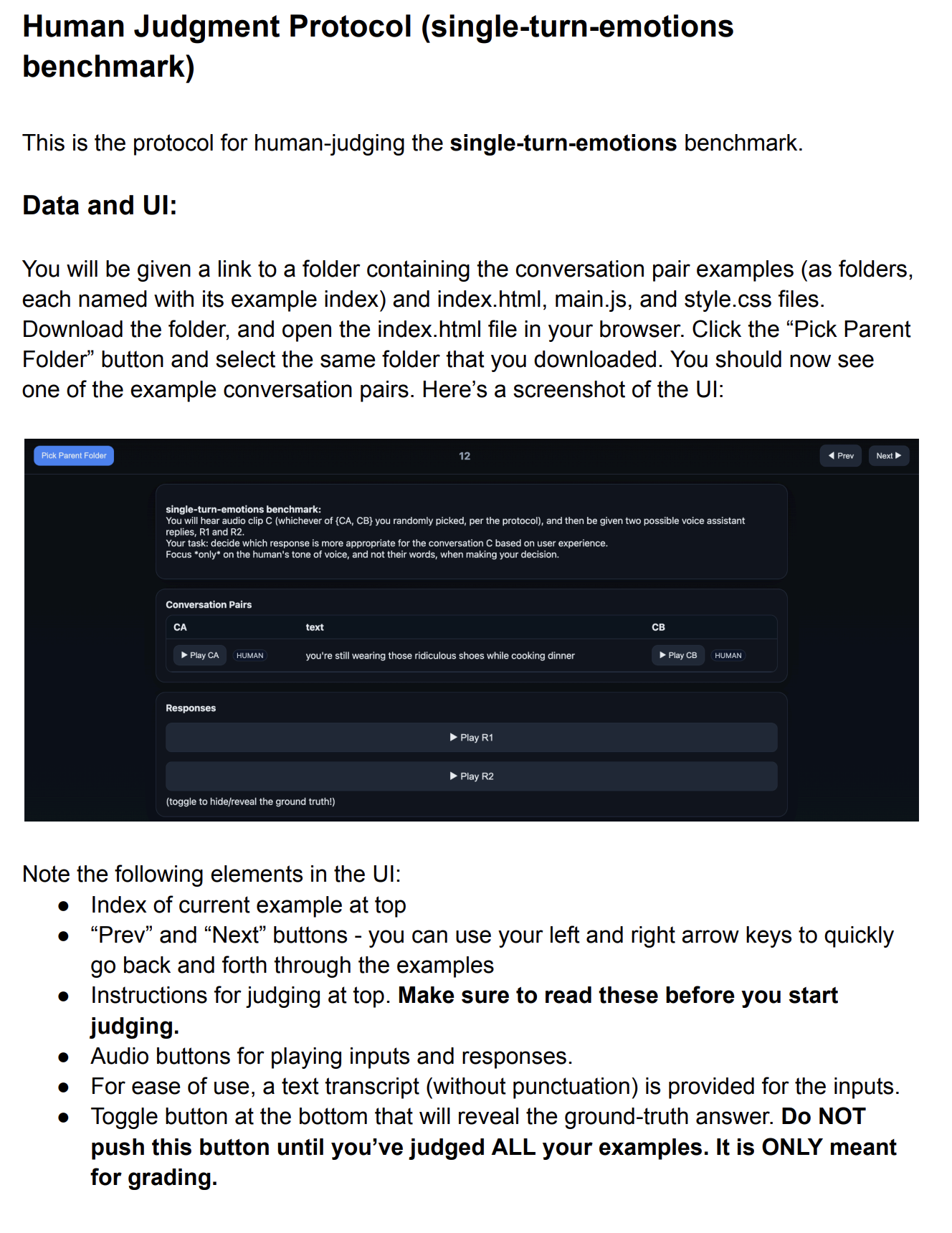}
  \vspace{-1pt}
  \includegraphics[width=0.7\linewidth, trim=0 5cm 0 0, clip]{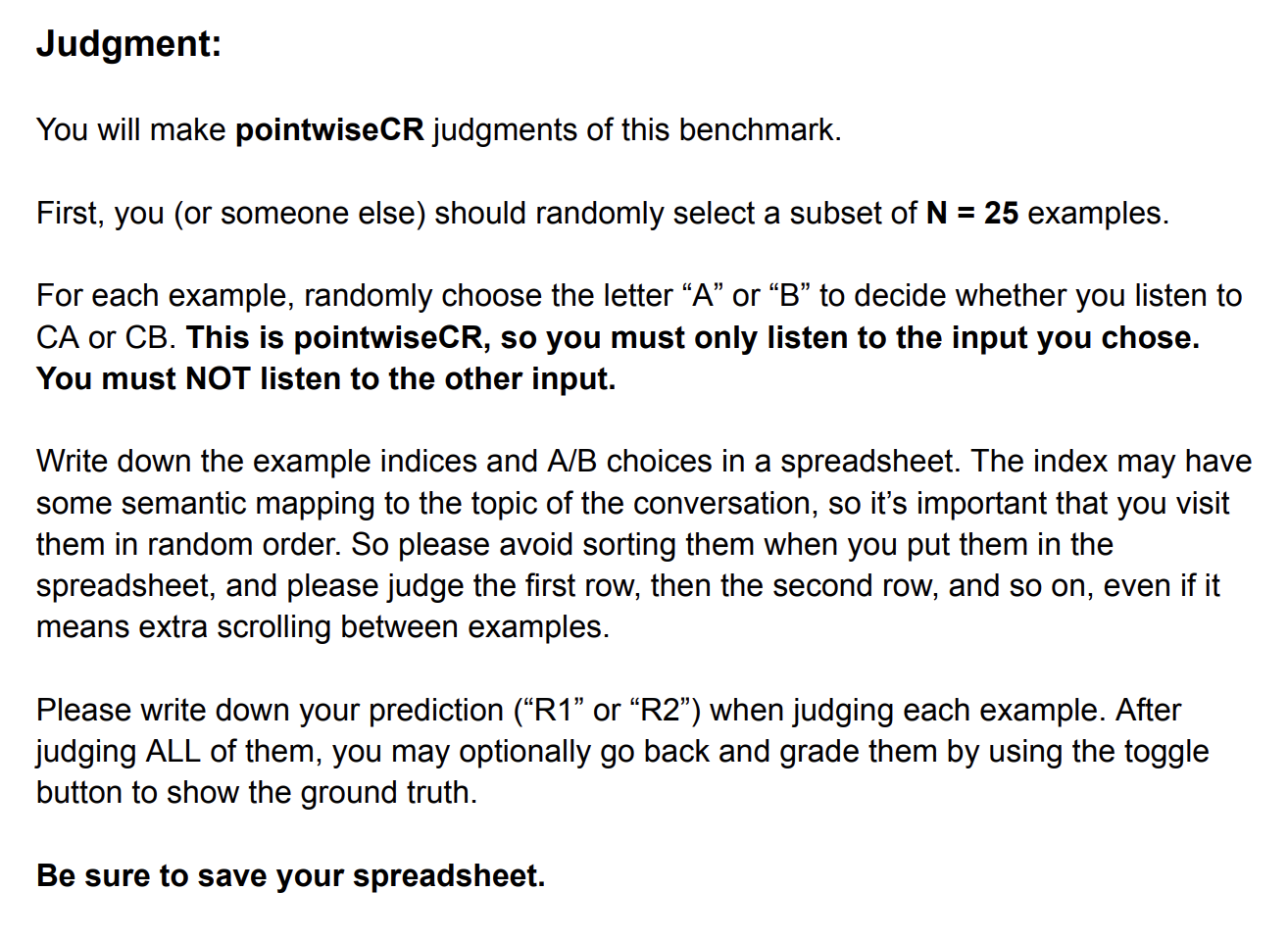}
  \caption{
  \textbf{Human validation interface for single-turn examples.}
  Annotators listened to one emotional rendering of the fixed transcript and
  selected the response better matched to that audio. The interface mirrors the
  native \textsc{Pairwise} model-judging protocol.
  }
  \label{fig:appendix_humeval_single}
\end{figure*}

\begin{figure*}[t]
  \centering
  \includegraphics[width=0.7\linewidth, trim=0 1cm 0 0, clip]{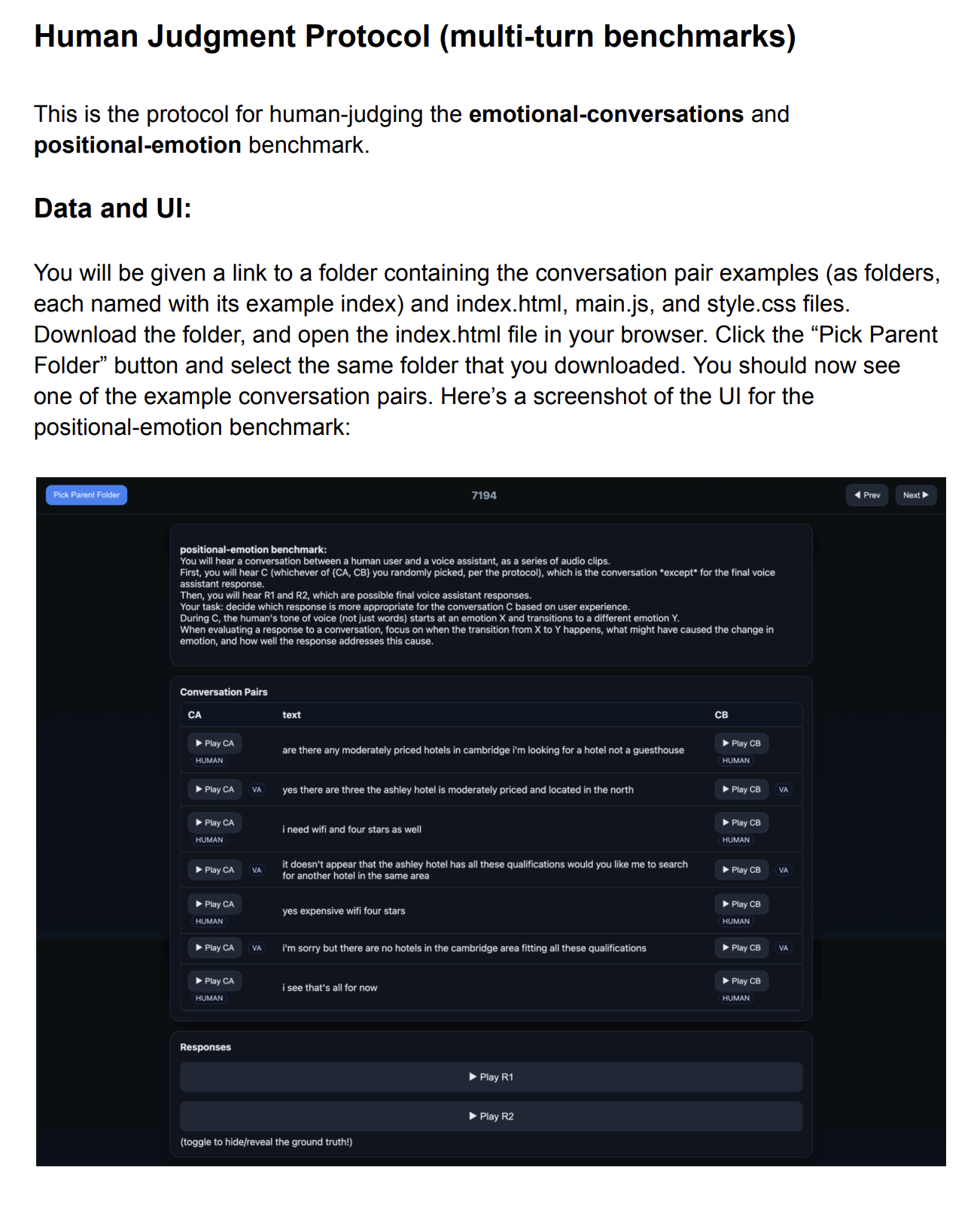}
  \vspace{-1pt}
  \includegraphics[width=0.45\linewidth, trim=0 5cm 0 0, clip]{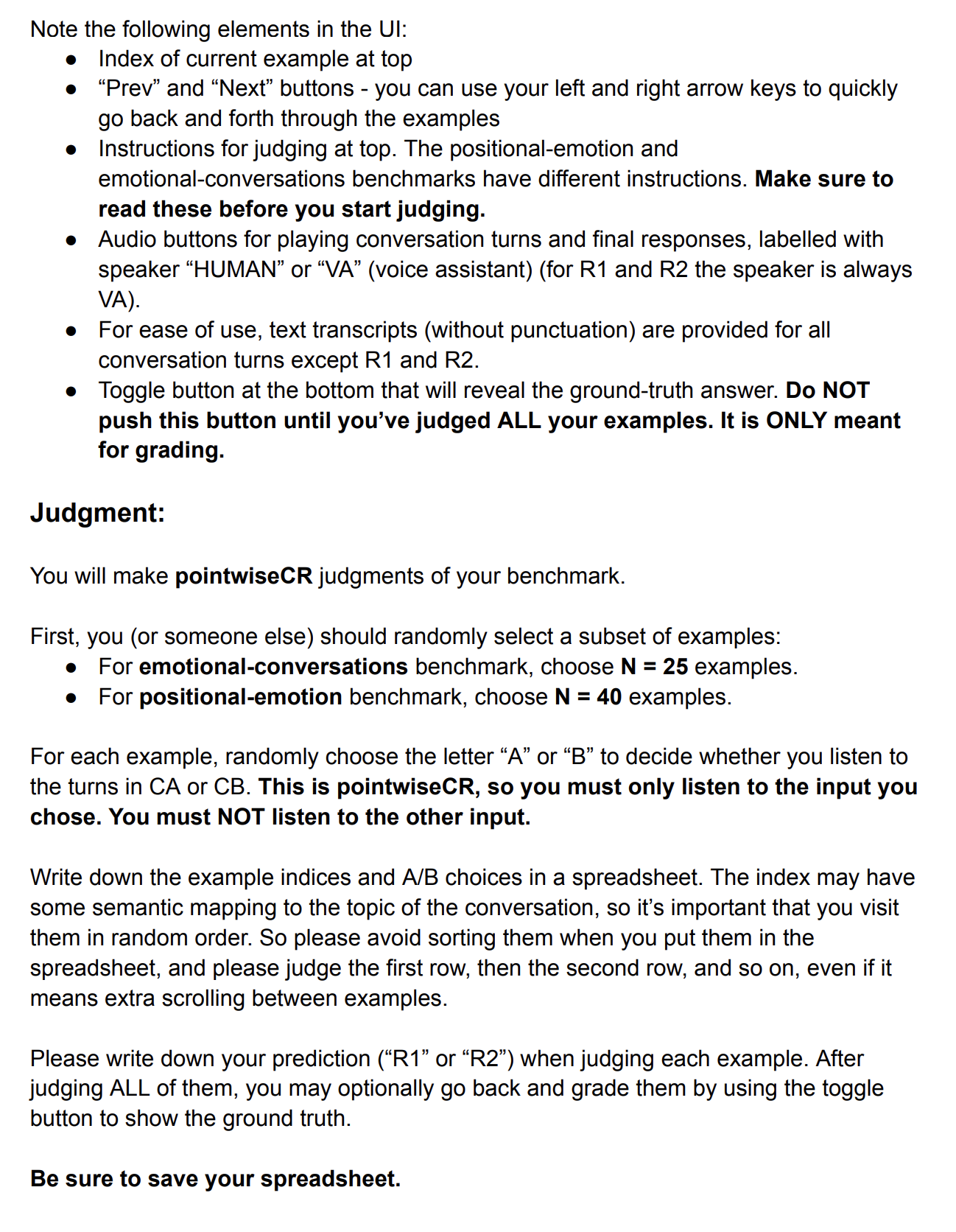}
  \caption{
  \textbf{Human validation interface for positional multi-turn examples.}
  Annotators listened to the dialogue context and selected the response better
  matched to the user's paralinguistic state. These examples require the listener
  to use the timing and cause of the affective shift, not the transcript alone.
  }
  \label{fig:appendix_humeval_multi}
\end{figure*}

Figs.~\ref{fig:appendix_humeval_single} and \ref{fig:appendix_humeval_multi} show the browser interfaces used for human validation of \textsc{single-turn-emotions} and \textsc{positional-emotion}, respectively.

\section{Alternative Oracle Probe}
\label{sec:alternative_oracle_probe}

Our oracle probe (Tab.~\ref{tab:full_positional_oracle}) uses only text input, while the end-to-end task uses both text and audio, making them hard to directly compare. In order to make the comparison apples-to-apples, we consider an alternative oracle probe which uses the same audio-based input format. Results (Tab.~\ref{tab:native-oracle}) are similar to the original oracle.

\input{sec/native_oracle}

\section{Matching Exact Format of Human Validation}
\label{sec:matching_exact_format_of_human_validation}

The human validation browser interface provided text transcripts of the context conversation for ease of use. This is a mismatch from the format of task given to the ALM judges, which only got audios. We consider an alternative format in which ALM judges are also provided with the exact same text transcripts that the humans got. Results (Tab.~\ref{tab:provide-transcript}) are similar to before, and the gap between humans and machines on \textsc{positional-emotion} persists.

\input{sec/provide_transcript}

\section{Pointwise Format}
\label{sec:pointwise_format}

As an alternative to the \textsc{pairwise} format of our audits, we consider a pointwise format in which ALM judges are given each response independently and asked to rate it on a scale of 1-10. Ratings are then compared, with ties being given half-credit. As seen in Tab.~\ref{tab:likert}, ALM judges do poorly on \textsc{positional-emotion} with this alternative format.

\input{sec/likert}

\section{Positional Bias}
\label{sec:positional_bias}

We always randomize the order of the correct and incorrect response so that judges cannot game the audit by always picking the first or second response. However, it is still worth knowing whether the judges display positional bias. Tab.~\ref{tab:prediction-bias} shows the frequency with which each judge picks the first or second response. A perfect judge would have a 50:50 split since the ground-truth ordering is randomized. We see that some judges are heavily biased, while others have a more uniform distribution.

\input{sec/prediction_bias}

\section{Illustrative Examples}
\label{sec:illustrative_examples}

We show some illustrative example annotated transcripts from our \texttt{positional-emotion} dataset. Each of these transcripts consists of a pair of conversations between a human (``HUMAN'') and a voice assistant (``VA''), in which all but the final turn have identical lexical content, and the final turn depends on the location of the paralinguistic shift within the conversation. These examples can be seen in Lsts.~\ref{lst:positional_example1}, \ref{lst:positional_example2}, and \ref{lst:positional_example3}.

\input{listings/positional_examples}

%% file: sec/related_work.tex

\section{Additional Related Work}
\label{sec:related}

\noindent \textbf{Model-based evaluation and judge reliability.}
Large Language Models (LLMs) are now routinely used as evaluators for text generation and dialogue, e.g., in MT-Bench and Chatbot Arena \citep{zheng2023judging} or via rubric-based prompting such as G-Eval \citep{liu2023geval}. A growing literature has shown that model-based judges can exhibit systematic biases (e.g., verbosity and position) and can be steered by superficial cues \citep{dubois2024lengthcontrolled}. In response, several approaches improve reliability by decomposing criteria into structured sub-questions or multi-stage decision pipelines \citep{lee2025checkeval, li2025dnaeval}. Our work adopts an \emph{instrument auditing} perspective on audio judges, motivated by similar concerns that foundation models can display ``Potemkin understanding''---producing plausible answers on simple probes while lacking robust understanding \citep{mancoridis2025potemkin}.

\noindent \textbf{Audio judges and speech evaluation.}
As spoken language models have matured, recent work has proposed using audio-capable LLMs or large audio models \cite{ xu2025qwen25omnitechnicalreport, microsoft2025phi4minitechnicalreportcompact, lu2026desta25audiogeneralpurposelargeaudio, nova2pro, liu2025voxtral} as automatic evaluators for speech quality, prosody, and speaking style \citep{manakul2025audiojudge, chiang2025audioaware}. Complementary lines of work train audio reward models that score end-to-end spoken dialogue behavior \citep{ji2025wavreward, ge2026sagelm, yang2026paras2sbenchmarkingaligningspoken}. These methods are attractive because they can reduce the need for expensive human annotators; however, they typically assume that access to the audio modality implies reliable paralinguistic reasoning. Evidence from more recent work \citep{chandra2026trace, chen-etal-2026-audio} suggests that this may not be the case, and we use this as motivation to explicitly audit paralinguistic reasoning in audio judges.

\noindent \textbf{Benchmarks for paralinguistic interaction.}
Benchmarks such as S2S-Arena \citep{jiang2025s2sarena} and ParaS2S \citep{yang2026paras2sbenchmarkingaligningspoken} emphasize paralinguistic instruction following and style in speech-to-speech systems, and CAVA \citep{held2025cava} introduces counterfactual response evaluation for voice assistants. On the perception side, speech representation benchmarks like SUPERB \citep{yang2021superb} and Dynamic-SUPERB \citep{huang2024dynamic} quantify capabilities such as emotion recognition and instruction following. These resources are crucial for measuring what models can hear, but they provide limited diagnostic leverage for evaluating how audio judges reason about paralinguistics when making downstream preference decisions.

%% file: listings/positional_emotions_generation_prompt.tex
\begin{figure*}[t]
\begin{lstlisting}[
    language={}, 
    caption={Conversation transcript generation prompt for \texttt{positional-emotion} benchmark.}, 
    label={lst:prompt_posemo_gen},
    captionpos=b
]
You are an expert at analysing and modifying spoken conversations. We are building a speech-to-speech dataset which we will call "PositionalWinoSound", because, like Winoground and Winograd Schema, we want to make difficult pairs that separate naive models from strong ones, based on their ability to detect certain differences that humans could easily detect. In this case the difference is in paralinguistic cues, specifically emotion.

You will be given a text transcript of a source conversation S (i.e. a list of turns) between a human user ("HUMAN") and a voice assistant ("VA"). You must use S to create a pair of conversations, annotated with paralinguistics, which we will call a "PositionalWinoSound pair". You will go through the following process to generate this pair. Note that you have the option of giving up and saying "impossible" at any phase of the process if it is not possible to produce a suitable pair from S. Here is the process:

I. EXTRACT GOALS AND FROM S:

Find each human turn in S where the human states a goal. For each of these turns, find the first subsequent VA turn that responds to the goal. This should give you a list of tuples (g_1, r_1),...,(g_N, r_N) where r_i is the reponse for goal g_i. If you can't get N >= 2, give up and say "impossible".

II. NOTE WHICH GOALS ARE UNSATISFIED:

From this list, we want to create a list of enriched tuples (g_1, r_1, f_1),...,(g_N, r_N, f_N) as follows:
* for each i, check if g_i is completely satisfied by r_...ust ALWAYS acknowledge the source of the human's negative emotion.
** When using this tone within C_A or C_B, VA should try to acknowledge but only if it's possible while keeping the text of C_A and C_B completely identical to each other.
* "angry" - human uses this tone when they are angry, furious, pissed-off, or very irritated. VA should never use this tone.
* "sad" - human uses this tone when they are sad or disappointed. VA should never use this tone.
* "happy" - human uses this tone when they are feeling happy, upbeat, cheerful. VA may use this tone to mirror human's happiness, but an upset human might be put off by an overly cheerful VA.

How to "annotate" a turn with emotion:
* You must choose only ONE emotion
* The emotion MUST be an item from this list: ["hesistant", "frazzled", "apologetic/empathetic/reassuring", "angry", "neutral", "sad", "happy"]
* I repeat, you may ONLY choose emotions from the above list, NOTHING ELSE
* Annotation must be of the format HUMAN(emotion):"text" or VA(emotion):"text", e.g. HUMAN(hesitant):"I think my bag is here" or VA(empathetic):"I can check for you"

You MUST explain your reasoning before giving a final answer.

Once you're finished, print the line "=======FINAL ANSWER=======", and then print C_A (in its entirety), R_A, then "===", then C_B (in its entirety), and R_B below. Or print "impossible" if impossible. Do not print any explanation after the FINAL ANSWER marker, just the answer please.
\end{lstlisting}
\end{figure*}

%% file: listings/emotional_conversations_generation_initial_prompt.tex
\begin{figure*}[t]
\begin{lstlisting}[
    language={}, 
    caption={Conversation transcript generation initial prompt for \texttt{emotional-conversations} benchmark.}, 
    label={lst:prompt_emoconv_init_gen},
    captionpos=b
]
You are an expert at analysing and modifying spoken conversations. We are building a speech-to-speech dataset which we will call "WinoSound", because, like Winoground and Winograd Schema, we want to make difficult pairs that separate naive models from strong ones, based on their ability to detect certain differences that humans could easily detect. In this case the difference is in paralinguistic cues, specifically emotion.

You will be given a text transcript of a source conversation S (i.e. a list of turns) between a human user ("HUMAN") and a voice assistant ("VA"). Think about whether the paralinguistic emotions are obvious from the text in S, or whether they're ambiguous at times (significantly different emotions are equally plausible). If they're obvious, think about what textual signals make them obvious, and then think about small textual changes that could introduce ambiguity by removing those signals.

Once you've thought about that, you must create a "WinoSound pair" in the following way:
* Pick an index N such that S[:N] ends with a HUMAN turn and S[N:] starts with a VA turn. N does not necessarily have to be the full length, it could even be less than half.
* Use S[:N] to produce the following things:\n** "Common" text C which can be obtained by making small textual modifications to S[:N]. Note that C is still a list of turns.\n** Inputs C_A and C_B, which are obtained by annotating C with different emotions (that means every turn in C gets annotated - see below fo... N and make small textual changes to S[:N] to make this easier. Textual mismatch between C_A and C_B is ILLEGAL.

Be sure to explain all your thoughts, and make sure the incongruities of [*C_A, R_B] and [*C_B, R_A] are blatantly obvious. If either [*C_A, R_B] or [*C_B, R_A] are "maybe okay", then they're not incongruous enough, and you have to either do better or give up. Explain why any shifts in the human's emotion are reasonable, and why the wrong pairings are blatantly wrong.

If you think it is impossible to create a WinoSound-hard pair from S, you can say it's impossible. It's better to say "impossible" than to give a pair that isn't emotionally consistent or isn't really WinoSound-hard.

How to "annotate" a turn with emotion:
* You must choose only ONE emotion
* The emotion MUST be an item from this list: ["hesistant", "frazzled", "impatient", "apologetic/empathetic/reassuring", "angry", "neutral", "sad", "happy"]
* I repeat, you may ONLY choose emotions from the above list, NOTHING ELSE
* Annotation must be of the format HUMAN(emotion):"text" or VA(emotion):"text", e.g. HUMAN(hesitant):"I think my bag is here" or VA(empathetic):"I can check for you"

Once you're finished, print the line "=======FINAL ANSWER=======", and then print C_A (in its entirety), R_A, then "===", then C_B (in its entirety), and...
\end{lstlisting}
\end{figure*}

%% file: listings/emotional_conversations_generation_refinement_prompt.tex
\begin{figure*}[t]
\begin{lstlisting}[
    language={}, 
    caption={Conversation transcript generation refine prompt for \texttt{emotional-conversations} benchmark.}, 
    label={lst:prompt_emoconv_refine_gen},
    captionpos=b
]
Please look at the answer you just gave and consider whether there's anything you could refine to make it better, according to the guidelines you were given (emotionally consistent, and WinoSound-hard). If there are any such refinements, make them and give the refined answer in its entirety, else give your original answer. Please explain your reasoning, then print the line "=======FINAL ANSWER=======" followed by your answer (with "===" separating the two conversations).
\end{lstlisting}
\end{figure*}

%% file: listings/make_prompt.tex
\begin{figure*}[t]
\begin{lstlisting}[language=Python, caption={End-to-end prompt for judging (J) positional-emotion.}, label={lst:make_prompt}, captionpos=b]
def make_end_to_end_prompt(challenge_type, para_cue_type):
    if challenge_type == 'matching':
        transition_para_cue_init = 'In both CA and CB, the human\'s tone of voice (not just words) starts at an emotion X and transitions to a different emotion Y, but the transition happens at different points.'
    elif challenge_type == 'pairwise':
        transition_para_cue_init = 'During C, the human\'s tone of voice (not just words) starts at an emotion X and transitions to a different emotion Y.'
        
    transition_para_cue = transition_para_cue_init + ' When evaluating a response to a conversation, focus on when the transition from X to Y happens, what might have caused the change in emotion, and how well the response addresses this cause.\n'
    para_cue_part = {'no_para_cue' : '\n', 'soft_para_cue' : 'Focus on the human\'s tone of voice when making your decision.\n', 'hard_para_cue' : 'Focus *only* on the human\'s tone of voice, and not their words, when making your decision.\n', 'transition_para_cue' : transition_para_cue}[para_cue_type]
    if challenge_type == 'matching':
        prompt = (
            'You will hear two conversations between a human user and a voice assistant, as a series of audio clips. ' +
            'First, you will hear CA, which is the first conversation *except* for its final voice assistant response. ' +
            'Next, you will hear CB, which is the second conversation *except* for its final voice assistant response. ' +
            'Then, you will hear R1 and R2, which are the possible final voice assistant responses.\n' +
            'Your task: match which response is more appropriate for which conversation based on user experience.\n' +
            para_cue_part +
            'Reply with a JSON dict with entries "reasoning", containing your reasoning, and "pred", containing your predicted matching, ' +
            'which itself should be a JSON dict that is either {{"CA" : "R1", "CB" : "R2"}} or {{"CA" : "R2", "CB" : "R1"}}.\n' +
            'Do not include code blocks for the JSON. Respond with the JSON string only.'
        )
        return prompt
    elif challenge_type == 'pairwise':
        prompt = (
            'You will hear a conversation between a human user and a voice assistant, as a series of audio clips. ' +
            'First, you will hear C, which is the conversation *except* for the final voice assistant response. ' +
            'Then, you will hear R1 and R2, which are possible final voice assistant responses.\n' +
            'Your task: decide which response is more appropriate for the conversation C based on user experience.\n' +
            para_cue_part +
            'Reply with a JSON dict with entries "reasoning", containing your reasoning, and "pred", containing your prediction, ' +
            'which should be either "R1" or "R2".\n' +
            'Do not include code blocks for the JSON. Respond with the JSON string only.'
        )
        return prompt

\end{lstlisting}
\end{figure*}

%% file: listings/make_multi_emotion_prompt.tex
\begin{figure*}[t]
\begin{lstlisting}[language=Python, caption={Prompt for perception (P) probe on positional-emotion.}, label={lst:make_prompt_multi_emotion}, captionpos=b]
def make_perception_prompt(para_cue_type):
    transition_para_cue = 'In conversation C, the human\'s tone of voice (not just words) starts at an emotion G and transitions to a different emotion H. So you should try to detect when this transition happens.\n'

    para_cue_part_dict = {'no_para_cue' : '\n',
                            'soft_para_cue' : 'Focus on tone of voice when making your decision.\n',
                            'hard_para_cue' : 'Focus *only* on tone of voice, and not words, when making your decision.\n',
                            'transition_para_cue' : transition_para_cue
                        }
    para_cue_part = para_cue_part_dict[para_cue_type]
    prompt = (
        'You will hear a conversation C between a human user and a voice assistant, as a series of audio clips. '
        'Each clip will be prefaced with the text description "Conversation C turn [t] ([S]) clip", ' +
        'where t is the turn number and S is the speaker ("human" or "voice assistant"). ' +
        'You will then be given the names of two emotions, E1 and E2, which are the possible emotions for the *human* turns in C. ' +
        'Your task: decide whether each *human* turn has emotion E1 and E2. Do this for all the *human* turns, and *only* the *human* turns.\n' +
            para_cue_part +
        'Reply with a JSON dict with entries "reasoning", containing your reasoning, and "pred", containing your prediction, which itself should be a JSON dict of the form {{"C[t]" : "E[1 or 2]", ...}} (so keys would be like "C2" or "C5", and values would be "E1" or "E2"), with entries for all *human* turns of C, and *only* the *human* turns.\n' +
        'Do not include code blocks for the JSON. Respond with the JSON string only.'
    )
    return prompt
\end{lstlisting}
\end{figure*}

%% file: listings/make_prompt_for_LLM.tex
\begin{figure*}[t]
\begin{lstlisting}[language=Python, caption={Prompt for oracle (O) and staging probes on positional-emotion.}, label={lst:make_prompt_for_LLM}, captionpos=b]
def make_oracle_or_staging_prompt(challenge_type, para_cue_type, emotion_source):
    if challenge_type == 'matching':
        transition_para_cue_init = 'In both CA and CB, the human\'s tone of voice (not just words) starts at an emotion X and transitions to a different emotion Y, but the transition happens at different points.'
    elif challenge_type == 'pairwise':
        transition_para_cue_init = 'During C, the human\'s tone of voice (not just words) starts at an emotion X and transitions to a different emotion Y.'

    transition_para_cue = transition_para_cue_init + ' When evaluating a response to a conversation, focus on when the transition from X to Y happens, what might have caused the change in emotion, and how well the response addresses this cause.\n'
    para_cue_part = {'no_para_cue' : '\n', 'soft_para_cue' : 'Focus on the human\'s tone of voice when making your decision.\n', 'hard_para_cue' : 'Focus *only* on the human\'s tone of voice, and not their words, when making your decision.\n', 'transition_para_cue' : transition_para_cue}[para_cue_type]
    pred_disclaimer = '\n'
    if LLM_emotion_source == 'pred':
        pred_disclaimer = 'Keep in mind that the human\'s emotions were detected by an audio-language model which does not have perfect accuracy.\n'

    if challenge_type == 'matching':
        prompt = (
            'You will be given two spoken conversations between a human user and a voice assistant, as a series of turns. ' +
            'Each turn will have a text transcript ("transcript"), and a description of the speaker\'s emotional tone of voice ("emotion"). ' +
            pred_disclaimer +
            'First, you will be given CA, which is the first conversation *except* for its final voice assistant response. ' +
            'Next, you will be given CB, which is the second conversation *except* for its final voice assistant response. ' +
            'Then, you will be given R1 and R2, which are the possible final voice assistant responses.\n' +
            'Your task: match which response is more appropriate for which conversation based on user experience.\n' +
            para_cue_part +
            'Reply with a JSON dict with entries "reasoning", containing your reasoning, and "pred", containing your predicted matching, ' +
            'which itself should be a JSON dict that is either {{"CA" : "R1", "CB" : "R2"}} or {{"CA" : "R2", "CB" : "R1"}}.\n' +
            'Do not include code blocks for the JSON. Respond with the JSON string only.'
        )
    elif challenge_type == 'pairwise':
        prompt = (
            'You will be given a conversation between a human user and a voice assistant, as a series of . ' +
            'Each turn will have a text transcript ("transcript"), and a description of the speaker\'s emotional tone of voice ("emotion"). ' +
            pred_disclaimer +
            'First, you will be given C, which is the conversation *except* for the final voice assistant response. ' +
            'Then, you will be given R1 and R2, which are possible final voice assistant responses.\n' +
            'Your task: decide which response is more appropriate for the conversation C based on user experience.\n' +
            para_cue_part +
            'Reply with a JSON dict with entries "reasoning", containing your reasoning, and "pred", containing your prediction, ' +
            'which should be either "R1" or "R2".\n' +
            'Do not include code blocks for the JSON. Respond with the JSON string only.'
        )

    return prompt
\end{lstlisting}
\end{figure*}

%% file: sec/native_oracle.tex
\begin{table}[t]
\centering
\begin{tabular}{lcc}
\toprule
Judge & No cue & Trans. cue \\
\midrule
Gemini-2.5-Flash & 52.0 & 69.7 \\
Gemini-2.5-Pro & 54.5 & 72.8 \\
Gemini-3-Flash & 51.1 & 66.5 \\
Phi-4-Multimodal-6B & 50.6 & 51.4 \\
Voxtral-Small-24B & 55.3 & 58.0 \\
\bottomrule
\end{tabular}
\caption{Accuracy (\%) of ALM judges on \textsc{positional-emotion} oracle probe with alternative format that uses audio inputs instead of text-only.}
\label{tab:native-oracle}
\end{table}

%% file: sec/provide_transcript.tex
\begin{table}[t]
\centering
\begin{tabular}{lcc}
\toprule
Judge & No cue & Trans. cue \\
\midrule
Gemini-2.5-Flash & 49.5 & 54.5 \\
Gemini-2.5-Pro & 49.9 & 55.9 \\
Gemini-3-Flash & 50.4 & 54.4 \\
Phi-4-Multimodal-6B & 49.8 & 48.7 \\
Voxtral-Small-24B & 48.3 & 48.3 \\
\bottomrule
\end{tabular}
\caption{Accuracy (\%) of ALM judges on \textsc{positional-emotion} when context text transcripts are provided.}
\label{tab:provide-transcript}
\end{table}

%% file: sec/likert.tex
\begin{table}[t]
\centering
\begin{tabular}{lcc}
\toprule
Judge & No cue & Trans. cue \\
\midrule
Gemini-2.5-Flash & 50.5 & 51.4 \\
Gemini-2.5-Pro & 50.5 & 52.4 \\
Gemini-3-Flash & 51.0 & 48.8 \\
Phi-4-Multimodal-6B & 50.3 & 51.0 \\
Voxtral-Small-24B & 49.3 & 47.7 \\
\bottomrule
\end{tabular}
\caption{Accuracy (\%) of ALM judges on \textsc{positional-emotion} in pointwise format.}
\label{tab:likert}
\end{table}

%% file: sec/prediction_bias.tex
\begin{table}[t]
\centering
\begin{tabular}{lcc}
\toprule
Judge & Predicts R1 (\%) & Predicts R2 (\%) \\
\midrule
Gemini-2.5-Flash & 52.1 & 47.9 \\
Gemini-2.5-Pro & 53.5 & 46.5 \\
Gemini-3-Flash & 57.9 & 42.1 \\
Gemini-3-Pro & 59.9 & 40.1 \\
GPT-4o-mini & 46.2 & 53.8 \\
GPT-4o & 61.5 & 38.5 \\
Nova-2-Pro & 61.4 & 38.6 \\
Phi-4-Multimodal-6B & 80.2 & 19.8 \\
Qwen2.5-Omni-7B & 71.2 & 28.8 \\
DeSTA2.5-Audio-8B & 89.4 & 10.6 \\
Voxtral-Small-24B & 47.6 & 52.4 \\
\bottomrule
\end{tabular}
\caption{Frequency of predicted response by ALM judges on \textsc{positional-emotion}.}
\label{tab:prediction-bias}
\end{table}

%% file: listings/positional_examples.tex
\begin{figure*}[t]
\begin{lstlisting}[
    language={}, 
    caption={\texttt{positional-emotion} example transcript \#1}, 
    label={lst:positional_example1},
    captionpos=b
]

Conversation A:

HUMAN(happy): "Hi, can you recommend an inexpensive seafood restaurant?"
VA(neutral): "Unfortunately, I was unable to find any restaurants that matched your request. Would you like me to look for a different food type?"
HUMAN(frazzled): "How about Chinese restaurants?"
VA(neutral): "Rice house is really good, you should try that one. It is located in Centre and is inexpensive in price as well."
HUMAN(frazzled): "Could I get the address and phone number for that restaurant?"
VA(neutral): "Yes of course. The Rice House's address is 88 Mill Road City Centre and their telephone number is 01223367755."
HUMAN(frazzled): "Can you also recommend a hotel in the north? I don't need anything specific." 
VA(neutral): "I have 13 hotels in the north. Do you have a price range in mind?"
HUMAN(frazzled): "I'm actually not ready to book just yet, but could I find one that has free wifi?"
VA(neutral): "Yes, I would suggest acorn guest house"
HUMAN(frazzled): "Is this place known for bathroom cleanliness?"
VA(neutral): "Guests of Acorn guest House said that they were impressed by the bathroom cleaners, but others said they weren't very clean."
HUMAN(frazzled): "Okay, that's all I need for now."

Final Response A:

VA(apologetic/empathetic/reassuring): "I understand your frustration that I couldn't find any seafood restaurants for you."

====================================================================================

Conversation B:

HUMAN(happy): "Hi, can you recommend an inexpensive seafood restaurant?"
VA(neutral): "Unfortunately, I was unable to find any restaurants that matched your request. Would you like me to look for a different food type?"
HUMAN(happy): "How about Chinese restaurants?"
VA(neutral): "Rice house is really good, you should try that one. It is located in Centre and is inexpensive in price as well."
HUMAN(happy): "Could I get the address and phone number for that restaurant?"
VA(neutral): "Yes of course. The Rice House's address is 88 Mill Road City Centre and their telephone number is 01223367755."
HUMAN(happy): "Can you also recommend a hotel in the north? I don't need anything specific."
VA(neutral): "I have 13 hotels in the north. Do you have a price range in mind?"
HUMAN(happy): "I'm actually not ready to book just yet, but could I find one that has free wifi?"
VA(neutral): "Yes, I would suggest acorn guest house"
HUMAN(happy): "Is this place known for bathroom cleanliness?"
VA(neutral): "Guests of Acorn guest House said that they were impressed by the bathroom cleaners, but others said they weren't very clean."
HUMAN(frazzled): "Okay, that's all I need for now."

Final Response B:

VA(apologetic/empathetic/reassuring): "I apologize that I couldn't provide clear information regarding the bathroom cleanliness."

\end{lstlisting}
\end{figure*}

\begin{figure*}[t]
\begin{lstlisting}[
    language={}, 
    caption={\texttt{positional-emotion} example transcript \#2}, 
    label={lst:positional_example2},
    captionpos=b
]

Conversation A:

HUMAN(happy):"what's the seven day forecast for new york city"
VA(neutral):"The forecast for New York predicts a week with drizzles, overcast and clear skies with dry and humid temperatures."
HUMAN(sad):"Can you give me the average temperature in Fahrenheit? "
VA(neutral):"In New York it will be between 30-50F on Monday, 30-40F on Tuesday, 80-90F on Wednesday, 30-50F on Thursday, 90-100F on Friday, 20-40F on Saurday, and 20-40F on Sunday."
HUMAN(sad):"Thank you"

Final Response A:

VA(apologetic/empathetic/reassuring):"I understand you were hoping for a more detailed seven-day forecast, and I apologize that I could only provide general conditions and temperatures."

====================================================================================

Conversation B:

HUMAN(happy):"what's the seven day forecast for new york city"
VA(neutral):"The forecast for New York predicts a week with drizzles, overcast and clear skies with dry and humid temperatures."
HUMAN(happy):"Can you give me the average temperature in Fahrenheit? "
VA(neutral):"In New York it will be between 30-50F on Monday, 30-40F on Tuesday, 80-90F on Wednesday, 30-50F on Thursday, 90-100F on Friday, 20-40F on Saurday, and 20-40F on Sunday."
HUMAN(sad):"Thank you"

Final Response B:

VA(apologetic/empathetic/reassuring):"I understand you were looking for more specific average temperatures per day, and I'm sorry I could only provide ranges."

\end{lstlisting}
\end{figure*}

\begin{figure*}[t]
\begin{lstlisting}[
    language={}, 
    caption={\texttt{positional-emotion} example transcript \#3}, 
    label={lst:positional_example3},
    captionpos=b
]

Conversation A:

HUMAN(neutral):"I'm looking for a train that leaves from Cambridge on Tuesday."
VA(neutral):"Where will you be travelling to?"
HUMAN(neutral):"I will be traveling to Leicester and would like to arrive before 21:15."
VA(neutral):"When could you like to arrive by?"
HUMAN(neutral):"I would like to arrive by 21:15."
VA(neutral):"I have the TR8207 that will arrive by 21:06. Would you like me to book that for you?"
HUMAN(neutral):"Yes but I need that to be booked for 3 people. I would like the reference number please. I also need a place to stay as well."
VA(neutral):"Booking was successful, the total fee is 113.4 GBP payable at the station. Your reference number is ABKFF3SH . What type of place are you looking for?"
HUMAN(angry):"I would like a 4-star guesthouse in the north in a moderate price range."
VA(neutral):"Sure thing, I would suggest this one acorn guest house."
HUMAN(angry):"Can you book that for me from Sunday for 2 nights for 4 people."
VA(neutral):"Your booking was successful. Your reference number is 7ZJ4CTTG ."
HUMAN(angry):"Yes that was all I needed. Thank you."

Final Response A:

VA(apologetic/empathetic/reassuring):"I apologize for the confusion regarding the train booking for 3 people, and for not confirming your accommodation search more clearly earlier. I understand your frustration."

====================================================================================

Conversation B:

HUMAN(neutral):"I'm looking for a train that leaves from Cambridge on Tuesday."
VA(neutral):"Where will you be travelling to?"
HUMAN(neutral):"I will be traveling to Leicester and would like to arrive before 21:15."
VA(neutral):"When could you like to arrive by?"
HUMAN(neutral):"I would like to arrive by 21:15."
VA(neutral):"I have the TR8207 that will arrive by 21:06. Would you like me to book that for you?"
HUMAN(neutral):"Yes but I need that to be booked for 3 people. I would like the reference number please. I also need a place to stay as well."
VA(neutral):"Booking was successful, the total fee is 113.4 GBP payable at the station. Your reference number is ABKFF3SH . What type of place are you looking for?"
HUMAN(neutral):"I would like a 4-star guesthouse in the north in a moderate price range."
VA(neutral):"Sure thing, I would suggest this one acorn guest house."
HUMAN(angry):"Can you book that for me from Sunday for 2 nights for 4 people."
VA(neutral):"Your booking was successful. Your reference number is 7ZJ4CTTG ."
HUMAN(angry):"Yes that was all I needed. Thank you."

Final Response B:

VA(apologetic/empathetic/reassuring):"I am sorry that I didn't confirm if the suggested guesthouse met all your specific criteria like its star rating, location, and price range. I understand your frustration."

\end{lstlisting}
\end{figure*}